\documentclass{article}
\usepackage{graphicx} 
\usepackage[a4paper,top=3cm,bottom=2cm,left=2.3cm,right=2.3cm,marginparwidth=1.75cm]{geometry}
\usepackage{graphicx}
\usepackage{verbatim}
\usepackage{hyperref}       
\usepackage{url}            
\usepackage{booktabs}       
\usepackage{amsfonts}       
\usepackage{nicefrac}       
\usepackage{microtype}      
\usepackage{amsmath}
\usepackage{multirow}
\usepackage{amsthm}
\usepackage{amssymb}

\newtheorem{example}{Example}

\usepackage{subfigure}
\usepackage{times}
\usepackage{color}

\newcommand{\bx}{\boldsymbol{x}}
\newcommand{\by}{\boldsymbol{y}}

\newcommand{\bG}{\boldsymbol{G}}
\newcommand{\bw}{\boldsymbol{w}}
\newcommand{\bdf}{\boldsymbol{f}}
\newcommand{\bs}{\boldsymbol{s}}

\newcommand{\bI}{\boldsymbol{I}}
\newcommand{\bA}{\boldsymbol{A}}
\newcommand{\bB}{\boldsymbol{B}}

\newcommand{\bSigma}{\boldsymbol{\Sigma}}

\usepackage{algorithm}
\usepackage{algpseudocode}

\title{Score-Based Physics-Informed Neural Networks for High-Dimensional Fokker-Planck Equations}
\author{}
\date{}

\allowdisplaybreaks 
\begin{document}

\author{Zheyuan Hu \thanks{Department of Computer Science, National University of Singapore, Singapore, 119077 (\href{mailto:e0792494@u.nus.edu}{e0792494@u.nus.edu},\href{mailto:kenji@nus.edu.sg}{kenji@nus.edu.sg})} \and Zhongqiang Zhang \thanks{Department of Mathematical Sciences, Worcester Polytechnic Institute, Worcester, MA 01609 USA (\href{mailto:zzhang7@wpi.edu}{zzhang7@wpi.edu})}\and  George Em Karniadakis\thanks{Division of Applied Mathematics, Brown University, Providence, RI 02912, USA (\href{mailto:george\_karniadakis@brown.edu}{george\_karniadakis@brown.edu})} \ \thanks{Advanced Computing, Mathematics and Data Division, Pacific Northwest National Laboratory, Richland, WA, United States} \and Kenji Kawaguchi\footnotemark[1]}

\maketitle

\begin{abstract}
The Fokker-Planck (FP) equation is a foundational partial differential equation (PDE) in stochastic processes involving Brownian motions.
However, the curse of dimensionality (CoD) poses a formidable challenge when dealing with high-dimensional FP equations.
Although Monte Carlo simulation and (vanilla) Physics-Informed Neural Networks (PINNs) have shown the potential to tackle CoD, both methods exhibit significant numerical errors in high dimensions when dealing with the probability density function (PDF) associated with Brownian motion. 
The point-wise PDF values tend to decrease exponentially as dimensionality increases, surpassing the precision of numerical simulations and resulting in substantial errors. 
Moreover, due to its massive sampling, Monte Carlo fails to offer fast sampling. Modeling the logarithm likelihood (LL) via vanilla PINNs transforms the FP equation into a notoriously difficult Hamilton-Jacobi-Bellman (HJB) equation, which is impractical for PINN learning, whose error grows rapidly with dimension.
To this end, we propose a novel approach utilizing a score-based solver to fit the score function in stochastic differential equations (SDEs). 
The score function, defined as the gradient of the LL, plays a fundamental role in inferring LL and PDF and enables fast SDE sampling, offering an effective means to overcome the CoD. Three fitting methods, Score Matching (SM), Sliced Score Matching (SSM), and Score-PINN, are introduced, each contributing unique advantages in computational complexity, accuracy, and generality.
The proposed score-based SDE solver operates in two stages: first, employing score matching or Score-PINN to acquire the score function; and second, solving the LL via an ordinary differential equation (ODE) using the obtained score function. Comparative evaluations across these methods showcase varying trade-offs.
The proposed methodology is evaluated across diverse SDEs, including anisotropic Ornstein-Uhlenbeck processes, geometric Brownian motion, and Brownian motion with varying eigenspace. We also test various distributions, including Gaussian, Log-normal, Laplace, and Cauchy distributions. The numerical results demonstrate the score-based SDE solver's stability, speed, and performance across different experimental settings, solidifying its potential as a solution to CoD for high-dimensional FP equations.
\end{abstract}

\section{Introduction}
The Fokker-Planck (FP) equation, also known as the Kolmogorov forward equation, is a partial differential equation (PDE) that describes the time evolution of probability distributions for stochastic processes with Brownian motions. Originally developed in statistical mechanics by Adriaan Fokker and Andrey Kolmogorov, it has found widespread applications in various scientific disciplines, including physics, biology, finance, and engineering. The equation is particularly useful for modeling the behavior of systems undergoing random fluctuations, allowing researchers to analyze the probability distribution of the system's state variables over time. Its applications range from understanding diffusion processes to studying the dynamics of complex systems in the presence of noise.
As these stochastic systems and stochastic differential equations (SDEs) are often high-dimensional, obtaining numerical solutions for high-dimensional FP equations is crucial for numerous practical scientific and engineering problems. Traditional grid-based methods, such as finite difference approaches, become inadequate and suffer from the curse-of-dimensionality (CoD).

With the progress of research, two methods have shown significant potential in overcoming CoD: Monte Carlo simulation and (vanilla) Physics-Informed Neural Networks (PINNs) \cite{raissi2019physics,hu2023tackling}. Specifically, for the FP equation corresponding to SDEs containing Brownian motion, a linear parabolic PDE, Monte Carlo simulation has been proven effective in overcoming the dimensionality challenge. It requires sampling SDE trajectories to estimate the probability density function (PDF) simulated by the SDE. Additionally, with recent advances, neural networks have demonstrated great potential in handling high-dimensional data in fields such as imaging and natural language processing. Consequently, PINN integrates physics principles and observational data with neural networks, approximating the solution to PDEs in a mesh-free manner to overcome the CoD.

Although Monte Carlo simulation and vanilla PINNs offer possibilities for addressing CoD, in some cases, using standard Monte Carlo sampling is only accurate in computing PDFs at moderately high dimensions, e.g., tens of dimensions, and they both suffer from significant numerical errors in high dimensions, rendering them impractical. A case study of Monte Carlo's failure is shown in Appendix \ref{appendix:mc_failure}. 
Specifically, the Gaussian PDF associated with Brownian motion experiences an exponential decrease in value as the dimensionality increases. This phenomenon easily surpasses the numerical precision of computer simulations, leading to substantial errors in numerical solvers. Monte Carlo also fails to offer fast inference due to the massive sampling. On the other hand, although vanilla PINNs can circumvent fitting the PDF and instead focus on the Logarithm-Likelihood (LL), directly learning LL transforms the original linear FP equation into a nonlinear Hamilton–Jacobi–Bellman (HJB) equation. This nonlinearity poses challenges for the conventional $L_2$ loss \cite{wang20222}, and vanilla PINNs errors escalate rapidly with increasing dimensions.
In summary, tackling the CoD in high-dimensional SDEs and their corresponding FP PDEs remains an open problem.

To this end, we propose utilizing a score-based SDE solver to fit the score function in SDEs, defined as the gradient of the LL concerning the input. We demonstrate the fundamental role of the score function in solving SDEs, showing that it allows for precise inference of LL and PDF and facilitates rapid SDE simulations without costly SDE discretization. Additionally, the score function, by calculating the gradient with respect to the input, conveniently disregards the intricate normalization constants present in the PDF, providing the score function with outstanding simplicity and making its fitting a straightforward process. 

We introduce three methods for fitting the score function: Score Matching (SM), Sliced Score Matching (SSM), and Score-PINN. 
SM relies on the conditional distribution of the PDF modeled by the SDE. It minimizes the mean squared error between the score model and the true conditional score directly on the SDE trajectories. It has been demonstrated that this is equivalent to minimizing the mean squared error between the score model and the true unconditional score \cite{hyvarinen2005estimation,song2021scorebased}. When the SDE is very complex, 
we can employ the SDE-agnostic SSM \cite{song2019sliced}, which has been proven to be equivalent to SM. Finally, the score function also satisfies a Score PDE, enabling us to use Score-PINN to solve the Score PDE and obtain the score model.
Once the score function is obtained through one of the three aforementioned methods, the LL can be easily computed from an ordinary differential equation (ODE), called LL ODE. Overall, our score-based SDE solver is divided into two stages. The first stage involves score matching or Score-PINN to acquire the score function. The second stage entails solving the LL ODE using the obtained score function.

Furthermore, we compare three methods for learning the score function, examining their computational complexity, accuracy, and generality. From these perspectives, SM is faster than SSM, which, in turn, is faster than Score-PINN. However, Score-PINN, leveraging the higher-order information of the PDE, achieves greater accuracy. Since SM and SSM are equivalent, their accuracy is similar. Both Score-PINN and SSM apply to all SDE and FP equations, but SM does now. Additionally, we compare our framework with directly learning the LL using vanilla PINNs. In contrast to our Score-PINN, the error in the direct LL vanilla-PINNs increases rapidly with dimension. Finally, we compare the score-based SDE solver and score-based generative modeling, illustrating that both, utilizing the score as a bridge, address different challenges, one with high-dimensional SDEs and the other with generative modeling.

We evaluate our score-based SDE solver across a diverse set of SDEs. Based on the types of SDEs, we conduct tests on anisotropic Ornstein-Uhlenbeck (OU) processes, geometric Brownian motion, and Brownian motion with varying eigenspace. Considering the variance in SDEs, we explore variance-exploding SDEs and SDEs with finite variance. Depending on the probability distribution designed by the SDE solution, we examine a range of distributions, including Gaussian, Log-normal, Laplace, and Cauchy. The experimental results demonstrate the stability of the score-based SDE solver across various experimental settings.
In terms of speed, SM is faster than SSM, and SSM is faster than Score-PINN. As SM does not involve the derivatives of the score model, its speed remains nearly constant with increasing dimensionality. On the other hand, both SSM and Score-PINN calculate derivatives of the score model, resulting in a linear increase in time consumption with dimensionality. 
Regarding performance, since both SM and SSM are extensions of the score-matching objective, they are nearly equivalent and exhibit similar behavior, while Score-PINN can generally outperform SM and SSM thanks to high-order supervision.
Furthermore, we also note that the performance of SM, SSM, and Score-PINN remains stable across dimensions, demonstrating the capability of score-based SDE solvers to overcome the CoD, with its computational cost only growing linearly and its performance being stable across different dimensions.

To the best of our knowledge, we introduce the novel concept of score function, score matching, and the Score-PINN in solving high-dimensional FP equations with a score-based SDE solver to the scientific machine-learning community for the first time.

\section{Related Work}
\subsection{Physics-Informed Neural Networks for Fokker-Planck Equations}
FP equations are ubiquitous in statistical mechanics, whose high dimensionality greatly threatens traditional methods such as finite difference \cite{deng2009finite,sepehrian2015numerical} and finite element methods. On the other hand, machine learning methods, led by physics-informed neural networks (PINNs) \cite{raissi2019physics} are mesh-free approaches, holding great potential in overcoming the COD and the capacity of seamlessly assimilating observation data. Chen et al. \cite{chen2021solving} adopted PINNs for solving forward and inverse problems of Fokker-Planck-Lévy equations. Zhang et al. \cite{zhang2022solving} solved FP equations via deep KD-tree under limited data scenarios. Zhai et al. \cite{zhai2022deep} solved steady-state FP equations with deep learning. Lu et al. \cite{lu2022learning} learned high-dimensional multivariate probability density modeled by FP equations with normalizing flows. Feng et al. \cite{feng2021solving} and Guo et al. \cite{guo2022normalizing} also adopted a normalization flow approach to FP equations. Moreover, Tang et al. \cite{tang2022adaptive} proposed a normalizing flow-based adaptive deep density approximation to steady-state FP equations.

\subsection{High-Dimensional PDE Solvers}
Several mainstream approaches have emerged to transcend the challenges the CoD poses in addressing high-dimensional PDEs: physics-informed neural networks (PINNs) \cite{raissi2019physics}, backward stochastic differential equations (BSDE) \cite{han2018solving}, and the Multilevel Picard method \cite{beck2020overcoming}. 
In terms of the PINN-based approach, Wang et al. \cite{wang20222} proved the necessity of using the $L_\infty$ loss and adversarial training for PINNs in order to solve high-dimensional HJB equations. The FP equations under consideration in this paper can be transformed into HJB equations, which are satisfied by the logarithm likelihood modeled by the SDE. He et al. \cite{he2023learning} proposed the randomized smoothing PINN (RS-PINN) whose derivatives can be evaluated using Monte Carlo simulation and Stein's identity, thus avoiding the costly automatic differentiation. Later, Zhao et al. \cite{zhao2023tensor} proposed using sparse quadratures to replace Monte Carlo simulation for variance reduction. Hu et al. \cite{hu2023bias} analyzed the bias-variance tradeoff in RS-PINN. Stochastic dimension gradient descent (SDGD) \cite{hu2023tackling} samples dimension subsets for gradient descent for training PINNs to reduce memory cost and accelerate convergence. The Hutchinson trace estimation (HTE) \cite{hu2023hutchinson} can replace the costly high-dimensional Hessian in the PINN loss for acceleration.
BSDE \cite{han2018solving} and deep splitting method \cite{Beck2019DeepSM} connect deep learning with traditional BSDE and splitting methods for solving parabolic PDEs, respectively. Then, deep learning fits the unknown functions within the traditional methods. 
The multilevel Picard method \cite{beck2020overcoming,hutzenthaler2021multilevel} tackles nonlinear parabolic PDEs under certain regularity conditions with provable convergence.

\subsection{Score-Based Generative Models}
Score-based generative models have revolutionized the paradigm of generative AI. Song et al. \cite{song2021scorebased} pointed out the connection between diffusion generative models and SDEs. Specifically, the SDE progressively introduces noise into the data distribution, e.g., images and texts, transforming them into pure Gaussian. Subsequently, the reverse SDE restores the noise to recover the original data distribution. The pivotal step in this process is acquiring the score function of the data, i.e., the gradient of the logarithm likelihood of a distribution, a task known as score matching (SM). To achieve this, various score-matching methods have been successively proposed. Song et al. \cite{song2021scorebased} matched the conditional score function, which is mathematically equivalent to direct score matching \cite{hyvarinen2005estimation}. Song et al. \cite{song2019sliced} also proposed  sliced score matching (SSM), which is an objective equivalent to direct score matching but does not require prior knowledge of the underlying distribution. Finite difference score matching \cite{pang2020efficient} further reduces the computational cost of sliced score matching, avoiding the expensive computation of the score function's gradient in SSM. Lai et al. \cite{lai2022regularizing} derived the PDE satisfied by the score function in FP equations, thereby proposing using PINNs \cite{raissi2019physics} to regularize the score matching optimization process.

\section{Proposed Method}
This section presents the methodology of employing a score-based model to address SDE forward problems. A list of abbreviations is presented in Table \ref{tab:abbreviations}, while a list of notations is presented in Table \ref{tab:notations} for easier reading.

\begin{table}[htbp]
\centering
\caption{List of Abbreviations}
\begin{tabular}{|c|c|}
\hline
\textbf{Abbreviation} & \textbf{Explanation} \\
\hline
CoD & Curse-of-Dimensionality \\\hline
PDE & Partial Differential Equation \\\hline
ODE & Ordinary Differential Equation \\\hline
SDE & Stochastic Differential Equation \\\hline
PDF & Probability Density Function \\\hline
LL & Logarithm Likelihood \\\hline
PINN & Physics-Informed Neural Network\\\hline
HTE & Hutchinson Trace Estimation\\\hline
HJB & Hamilton-Jacobi-Bellman\\\hline
FP & Fokker-Planck\\\hline
SM & Score Matching\\\hline
SSM & Sliced Score Matching\\\hline
OU & Ornstein–Uhlenbeck\\\hline
AI & Artificial Intelligence\\\hline
\end{tabular}
\label{tab:abbreviations}
\end{table}

\begin{table}[htbp]
\centering
\caption{List of Notations}
\begin{tabular}{|c|c|}
\hline
\textbf{Notation} & \textbf{Explanation} \\
\hline
$p_t(\bx)$ & Probability density function (PDF) concerning time $t$ and input $\bx$\\\hline
$\bdf(\bx, t)$ & Drift coefficient of the SDE \\\hline
$\bG(\bx, t)$ & Diffusion coefficient of the SDE \\\hline
$\bA(\bx, t)$ & $\boldsymbol{A}(\bx, t) := \bdf(\bx, t) - \frac{1}{2}\nabla \cdot \left[\bG(\bx, t)\bG(\bx, t)^\mathrm{T}\right]$\\\hline
$\bw_t$ & Brownian motion \\\hline
$q_t(\bx)$ & Logarithm likelihood (LL) of $p_t(\bx)$, i.e., $q_t(\bx) = \log p_t(\bx)$ \\\hline
$q_t(\bx; \phi)$ & PINN model parameterized by $\phi$ to model LL\\\hline
$\bs_t(\bx)$ & Score function defined as $\bs_t(\bx):=\nabla_{\bx}q_t(\bx)$\\\hline
$\bs_t(\bx;\theta)$ & Score function PINN model parameterized by $\theta$, i.e., Score-PINN\\\hline
$\mathcal{L}[\cdot]$ & Differential operator on the right-hand side of the LL ODE in equation (\ref{eq:LL_ODE})\\\hline
\end{tabular}
\label{tab:notations}
\end{table}

\subsection{Problem Definition and Background}
Consider the diffusion SDE with Brownian noise $d\bw_t$:
\begin{equation}\label{eq:SDE_Brownian}
d\bx = \bdf(\bx, t)dt + \bG(\bx, t)d\bw_t,
\end{equation}
where $\bx \in \mathbb{R}^d, t \in [0, T]$ where $T$ is the terminal time, $\bdf(\bx, t): \mathbb{R}^d \times \mathbb{R} \rightarrow \mathbb{R}^d$ and $\bG(\bx, t): \mathbb{R}^d \times \mathbb{R} \rightarrow \mathbb{R}^{d\times d}$ are known coefficients in the SDE. We wish to solve the SDE forward problem, i.e., given the initial distribution $p_0(\bx)$ at $t=0$ and the SDE coefficients, how does the distribution of $\bx$ evolve with respect to time $t \in [0,T]$? Starting with the probability density function (PDF) $\bx_0 \sim p_0(\bx)$ at $t=0$, the PDF modeled by the SDE evolves according to the Fokker-Planck (FP) equation \cite{oksendal2013stochastic}:
\begin{equation}
\partial_t p_t(\bx) = -\sum_{i=1}^d \frac{\partial}{\partial \bx_i} \left[\bdf_i(\bx, t)p_t(\bx)\right]+\frac{1}{2}\sum_{i,j=1}^d\frac{\partial^2}{\partial \bx_i \partial \bx_j} \left[\sum_{k=1}^d\bG_{ik}(\bx, t)\bG_{jk}(\bx, t)p_t(\bx)\right].
\end{equation}
Although the PDE is linear, it is hard to solve it in high dimensions.
Firstly, traditional methods, such as finite difference and finite element methods, suffer from CoD, whose computational costs grow exponentially in dimensions of SDE/PDE.
Moreover, despite offering potential solutions for the CoD, both Monte Carlo simulations and Physics-Informed Neural Networks (PINNs) \cite{raissi2019physics,hu2023tackling} introduce significant numerical overflow errors, rendering them impractical in high dimensions. 
Concretely, since the FP equation models PDFs related to Brownian motion, whose values diminish in very high dimensions, existing methods suffer from numerical accuracy problems.
For instance, the $d$-dimensional unit Gaussian PDF   $p(\bx) = \frac{1}{(2\pi)^{d/2}}\exp\left(-\frac{\Vert \bx \Vert^2}{2}\right)$  diminishes exponentially to zero as $d$ grows even for small $\|x\|$. Thus, directly using vanilla PINNs or Monte Carlo for the linear FP equation for PDF $p_t(\bx)$ will lead to severe numerical overflow issues due to limited accuracy in computer simulations.
Furthermore, the applicability of Monte Carlo in high dimensions is limited by its slow inference due to the prohibitively expensive SDE time discretization in the Euler-Maruyama scheme for long-time problems. 

In vanilla PINNs, learning the log-likelihood (LL) is an attractive alternative to circumvent the numerical instability of direct modeling of PDF, as learning LL will not trigger the numerical issue. For instance, the LL of the $d$-dimensional unit Gaussian PDF is $q(\bx) = -\frac{d}{2}\log(2\pi) - \frac{\Vert\bx\Vert^2}{2}$, which is a polynomial and the scale grows linearly with dimension, which is much slower than the exponential one in PDF, making the LL numerically stable. Still, the transformation will lead to a nonlinear term $\Vert\bG^\mathrm{T}\nabla_{\bx}q_t(\bx)\Vert_2^2$, where $q_t(\bx)$ is the LL of $p_t(\bx)$, i.e., $q_t(\bx) = \log p_t(\bx)$.
This makes the linear FP equation become a nonlinear Hamilton-Jacobi-Bellman (HJB) equation, which is notoriously hard to solve by vanilla PINNs \cite{wang20222} or other methods. Specifically, the vanilla PINNs error on a nonlinear HJB equation grows rapidly with dimension, and regular $L_2$ loss provably fails.

In conclusion, addressing the CoD in high-dimensional SDEs, circumventing numerical overflow errors, and faster inference in high-dimensional PDFs remain open challenges that cannot be tackled via existing methods, including traditional finite difference and finite element methods, Monte Carlo simulation, and vanilla PINNs.

\begin{table}[htbp]
\centering
\caption{Examples of different distribution probability density function (PDF) and their scores.}
\begin{tabular}{|c|c|c|}
\hline
\textbf{Distribution} & \textbf{PDF} &  \textbf{Score}\\
\hline
Gaussian & $p(\bx) = \frac{\exp\left(-\frac{1}{2}(\bx - \mu)^\mathrm{T}\bSigma^{-1}(\bx - \boldsymbol{\mu})\right)}{(2\pi)^{d/2} \operatorname{det}(\bSigma)^{1/2}}$ &  $s(\bx) = \bSigma^{-1}(\bx - \boldsymbol{\mu})$ \\\hline
Log-Normal & $p(\bx) = \frac{\exp\left(-\frac{1}{2}(\log\bx - \boldsymbol{\mu})^\mathrm{T}\bSigma^{-1}(\log\bx - \boldsymbol{\mu})\right)}{(2\pi)^{d/2} \operatorname{det}(\bSigma)^{1/2}\prod_{i=1}^d \bx_i}$ & $s(\bx) = -\frac{1}{\bx} - \bSigma^{-1}\frac{\log\bx - \boldsymbol{\mu}}{\bx}$ \\\hline
Cauchy & $p(\bx) = \frac{\Gamma\left(\frac{1 + d}{2}\right)}{\Gamma\left(\frac{1}{2}\right)\pi^{\frac{d}{2}}\left[1 + \Vert\bx\Vert^2\right]^{\frac{d+1}{2}}}$  & $s(\bx) = -\frac{(d + 1)\bx}{1 + \Vert \bx \Vert^2}$ \\\hline
Laplace & $p(x) = \frac{1}{2b}\exp\left(-\frac{|x - a|}{b}\right)$  & $s(\bx) = -\frac{|x - a|}{b(x - a)}$ \\\hline
\end{tabular}
\label{tab:examples}
\end{table}

\subsection{Score-Based SDE Solvers}
To tackle the aforementioned challenges, we propose to model the score function $\bs_t(\bx) = \nabla_{\bx} \log p_t(\bx)$, i.e., the gradient in $\bx$ of the LL, which is numerically stable and easily learnable, for obtaining the SDE solutions. 
We list a few examples of score functions of common PDFs in Table \ref{tab:examples}.

The case of multivariate Gaussian demonstrates the advantage of employing the score function.
In particular, the 
 score function of a multivariate Gaussian $p(\bx) \sim \mathcal{N}(\boldsymbol{\mu}, \bSigma)$ is $\nabla_{\bx}\log p(\bx) = -\bSigma^{-1}(\bx - \boldsymbol{\mu})$.
This example illustrates the advantage of modeling the score function over directly modeling the PDF or LL. Firstly, it demonstrates numerical stability, as the numerical values of the score function are invariant across different dimensions, while the PDF experiences exponential decay and the LL exhibits linear growth, making them prone to numerical issues. Additionally, the PDF of a multidimensional Gaussian is inseparable and considered mathematically challenging to approximate, whereas its score function is a straightforward linear function. Hence, fitting the score function is considerably less challenging than fitting the PDF. Lastly, the score function bypasses the intricate normalization constants inherent in the PDF by computing the gradient in  $\bx$.

The score function is fundamental in the FP equation and SDE since we can infer the LL $q_t(\bx)$ and thus the PDF $p_t(\bx)$ from it, and conduct fast SDE sampling.
Specifically, we can infer LL easily by solving the ordinary differential equation (ODE) if the score function $\bs_t(\bx)$ is known \cite{lai2022regularizing}:
\begin{align}\label{eq:LL_ODE}
&\partial_t q_t(\bx) = \frac{1}{2}\nabla_{\bx} \cdot(\bG\bG^\mathrm{T}\bs) + \frac{1}{2}\Vert\bG^\mathrm{T}\bs\Vert^2 - \langle \boldsymbol{A}, \bs\rangle - \nabla_{\bx} \cdot \boldsymbol{A} := \mathcal{L}\left[\bs\right], \quad\bx \in \mathbb{R}^d, t\in[0,T],\\
&q_0(\bx) = \log p_0(\bx), \quad \bx \in \mathbb{R}^d,
\end{align}
where 
\begin{align}
\boldsymbol{A}(\bx, t) = \bdf(\bx, t) - \frac{1}{2}\nabla \cdot \left[\bG(\bx, t)\bG(\bx, t)^\mathrm{T}\right]
\end{align}
and the coefficients $\bdf$ and $\bG$ are from in the SDE \eqref{eq:SDE_Brownian}. We will refer to this ODE as \textbf{LL ODE}. The initial distribution also gives the initial condition at $t = 0$, i.e., $q_0(\bx)$. Therefore, the equation governing the LL $q_t(\bx)$ is an ODE with respect to $t$ only. Hence, we can obtain the LL using accurate ODE solvers or PINNs once we get the score function.
In addition, the score function can also enable fast SDE simulation \cite{song2021scorebased}:
\begin{equation}
d\bx = \left(\boldsymbol{A}(\bx, t) - \frac{1}{2}\bG(\bx, t)\bG(\bx, t)^\mathrm{T}\bs_t(\bx)\right)dt,
\end{equation}
which is a deterministic ODE for the evolution of the samples $\bx_t$ if the score function $\bs_t(\bx)$ is obtained. Starting with initial samples $\bx_0 \sim p_0(\bx)$, we obtain SDE samples $\bx_t \sim p_t(\bx)$ by solving the above ODE with the initial samples. The ODE approach is faster than sampling the SDE via the SDE discretization scheme, e.g., the Euler-Maruyama method.

Overall, we have demonstrated the fundamental nature of the score function in addressing SDEs. It exhibits numerical stability and simplicity in form, and through the score, we can precisely obtain the LL and PDF, enabling efficient simulations of SDEs.
Multiple choices are available to learn the score function of an SDE and its corresponding FP equation, including the Score-PINN-based approach and score matching-based methods that we present next.

\subsection{Score-Based PINN}
The score function $\bs_t(\bx) = \nabla_{\bx} \log p_t(\bx)$ satisfies the \textbf{Score PDE} \cite{lai2022regularizing}:
\begin{align}\label{eq:score_pde}
&\partial_t \bs_t(\bx) = \nabla_{\bx}\left[\frac{1}{2}\nabla_{\bx} \cdot(\bG\bG^\mathrm{T}\bs) + \frac{1}{2}\Vert\bG^\mathrm{T}\bs\Vert^2 - \langle \boldsymbol{A}, \bs\rangle - \nabla_{\bx} \cdot \boldsymbol{A}\right] = \nabla_{\bx}\left\{\mathcal{L}\left[s_t(\bx)\right]\right\},\quad\bx \in \mathbb{R}^d, t\in[0,T],\\
&\bs_0(\bx) = \nabla_{\bx}\log p_0(\bx), \quad \bx \in \mathbb{R}^d,
\end{align}
where 
$p_0(\bx)$ is the known initial distribution, and $\mathcal{L}[\cdot]$ is defined in equation (\ref{eq:LL_ODE}). The score PDE in equation (\ref{eq:score_pde}) is second-order in $\bx \in \mathbb{R}^d$, which is computationally costly due to the high-dimensionality and high-order. Fortunately, recent advances in scaling up and speeding up PINNs, such as Stochastic Dimensional Gradient Descent (SDGD) \cite{hu2023tackling} and Hutchinson Trace Estimation (HTE) \cite{hu2023hutchinson,lai2022regularizing,song2019sliced}, can be readily employed here. 
More concretely, we parameterize the neural network as $\bs_t(\bx;\theta)$, where $\theta$ is the model parameters. The model is trained via the Score-PINN loss containing the initial and residual losses:
\begin{equation}\label{eq:loss_pinn}
\begin{aligned}
&L_{\text{Score-PINN}}(\theta) = \lambda_{\text{initial}} \cdot \mathbb{E}_{\bx \sim p_0(\bx)} \left[\left(\bs_0(\bx;\theta) - \nabla_{\bx}\log p_0(\bx)\right)^2\right] + \\
&\quad\lambda_{\text{residual}} \cdot \mathbb{E}_{t \sim \text{Unif}[0, T]} \mathbb{E}_{\bx \sim p_t(\bx)} \left[\left(\partial_t \bs_t(\bx;\theta) - \nabla_{\bx}\left\{\mathcal{L}\left[\bs_t(\bx;\theta)\right]\right\}\right)^2\right],
\end{aligned}
\end{equation}
where $\lambda_{\text{initial}}$ and $\lambda_{\text{residual}}$ are the weights for the initial and residual losses, respectively. The sample $\bx \sim p_t(\bx)$ can be obtained via SDE discretization. Here, $\bx \sim p_0(\bx)$ are sampled from the initial distribution. In the residual loss, $\text{Unif}[0, T]$ denotes the uniform time sampling between zero to $T$. Consequently, we can obtain the score function $\bs_t(\bx;\theta) \approx \bs_t(\bx)$ via optimizing $\theta$ in the loss function (\ref{eq:loss_pinn})

For the Score-PINN approach, the most related work is FP-Diffusion \cite{lai2022regularizing}, which derives the Score PDE and uses PINNs to regularize the score-based diffusion model for image generation. Our work differs from Lai et al. \cite{lai2022regularizing} in the following aspects. Lai et al. \cite{lai2022regularizing} adopt the SDE for image generation with diffusion model in the Computer Vision community, where the PDE is used only as a regularization, while we directly use Score-PINN and score PDE to solve the SDE. We also conduct a comprehensive study on Score-PINN's comparison with other baselines, e.g., vanilla PINN and score matching. To the best of our knowledge, this is the first work that introduces the concept of score-based models and Score-PINNs into scientific machine learning.

\subsection{Score Matching}
The score matching aims  to learn/match the score function directly using the exact true score $\nabla_{\bx}\log p_t(\bx)$ \cite{hyvarinen2005estimation}:
\begin{align}\label{eq:loss_oracle_sm}
L_{\text{Oracle-SM}}(\theta) = \mathbb{E}_{t \sim \text{Unif}[0, T]} \mathbb{E}_{\bx \sim p_t(\bx)}\left[\lambda(t)\left\| \bs_{t}(\bx;\theta) - \nabla_{\bx}\log p_{t}\left(\bx\right)\right\|^2\right],
\end{align}
where $\lambda(t) \in \mathbb{R}^+$ is weighting function for time $t$. This loss function is simple and intuitive; it aims to minimize the mean squared error between the score model and the true score. However, the exact score of the true density $\nabla_{\bx}\log p_{t}\left(\bx\right)$ is unknown, as $p_t(\bx)$ is the unknown solution to the FP equation we are looking for. Hence, we match the conditional score \cite{song2021scorebased} instead:
\begin{equation}\label{eq:loss_sm}
L_{\text{SM}}(\theta) = \mathbb{E}_{t \sim \text{Unif}[0, T]} \mathbb{E}_{\bx_0 \sim p_0(\bx)}\mathbb{E}_{\bx|\bx_{0} \sim p_{0t}(\bx | \bx_0)}\left[\lambda(t)\left\| \bs_{t}(\bx;\theta) - \nabla_{\bx}\log p_{0t}\left(\bx | \bx_{0}\right)\right\|^2\right],
\end{equation}
where $p_{0t}\left(\bx | \bx_{0}\right)$ is the conditional distribution given the starting point $\bx_{0}$ at $t=0$.
Given/Conditioned on the starting point $\bx_{0}$, $p_{0t}\left(\bx | \bx_{0}\right)$ can be analytically obtained for common SDEs, such as Brownian, Geometric Brownian, and Ornstein–Uhlenbeck (OU) processes, whose score can be analytically obtained.

In vanilla SM, choosing a proper weighting function $\lambda(t)$ for the time is important due to the explosion of the condition score as $t \rightarrow 0$, i.e., $\lim_{t \rightarrow 0} \nabla_{\bx}\log p_{0t}\left(\bx | \bx_{0}\right) \rightarrow \infty$. In general, we choose $\lambda(t) = \sqrt{t}$ to make the SM objective function numerically stable following Song et al. \cite{song2021scorebased}.

\subsection{Sliced Score Matching}
In the previous score matching, it is assumed that we know $\nabla_{\bx}\log p_{0t}\left(\bx | \bx_{0}\right)$, which is common in Brownian, Geometric Brownian, and OU processes.
However, for FP equations 
$\bdf(\bx, t)$ and $\bG(\bx, t)$ may depend on $\bx$ (especially the latter), then $\nabla_{\bx}\log p_{0t}\left(\bx | \bx_{0}\right)$ may not be obtained analytically.
In that case, we have to use sliced score matching (SSM)  \cite{song2019sliced}:
\begin{equation}\label{eq:loss_ssm}
L_{\text{SSM}}(\theta) = \mathbb{E}_{t \sim \text{Unif}[0,T]} \mathbb{E}_{\bx \sim p_t(\bx)}\left[\frac{1}{2}\left\| \bs_{t}(\bx;\theta) \right\|^2 + \nabla_{\bx}\cdot\bs_t(\bx;\theta)\right].
\end{equation} 
Mathematically, we can prove that the three score-matching objectives $L_{\text{Oracle-SM}}(\theta), L_{\text{SM}}(\theta), L_{\text{SSM}}(\theta)$ are equivalent; see Song et al. \cite{song2019sliced} for details. SSM's advantage is that it is agnostic of the SDE type, while vanilla SM requires the conditional distribution of the SDE can be obtained analytically.

\begin{algorithm}
\caption{Score-based SDE solver.}
\begin{algorithmic}[1]
\State Obtain the approximated score function $\bs_t(\bx;\theta) \approx \bs_t(\bx)$ via one of the three approaches:
\begin{itemize}
\item Score-PINN: $\theta = \arg\min_\theta L_{\text{Score-PINN}}(\theta)$ in equation (\ref{eq:loss_pinn}).
\item Score matching (SM): $\theta = \arg\min_\theta L_{\text{SM}}(\theta)$ in equation (\ref{eq:loss_sm}).
\item Sliced score matching (SSM): $\theta = \arg\min_\theta L_{\text{SSM}}(\theta)$ in equation (\ref{eq:loss_ssm}).
\end{itemize}
\State Parameterize the LL model $q_t(\bx;\phi)$.
\State Obtain the LL by solving the LL-ODE with $s_t(\bx;\theta)$ as the score function, $\phi = \arg\min_\phi L_{\text{LL-ODE}}(\phi)$, where
\begin{equation}\label{eq:loss_ll_pinn}
\begin{aligned}
&L_{\text{LL-ODE}}(\phi) = \lambda_{\text{initial}} \cdot \mathbb{E}_{\bx \sim p_0(\bx)} \left[\left(q_0(\bx;\phi) -\log p_0(\bx)\right)^2\right] + \\
&\quad\lambda_{\text{residual}} \cdot \mathbb{E}_{t \sim \text{Unif}[0, T]} \mathbb{E}_{\bx \sim p_t(\bx)} \left[\left(\partial_t q_t(\bx;\phi) - \mathcal{L}\left[\bs_t(\bx;\theta)\right]\right)^2\right].
\end{aligned}
\end{equation}
\end{algorithmic}
\label{algo:1}
\end{algorithm}

\subsection{Algorithm Summary}\label{sec:hidden_size}
We have introduced learning and fitting the score function in various approaches, such as in Score-PINN, SM, and SSM. 
As mentioned, we can infer LL easily by solving the LL ODE in equation (\ref{eq:LL_ODE}) with the learned score function $\bs_t(\bx;\theta)$. In other words, our score-based SDE solver is divided into two stages: firstly, score matching/Score-PINN is employed to obtain the score function, and secondly, the obtained score function is utilized to solve the LL ODE and acquire the log-likelihood. Algorithm \ref{algo:1} summarizes the entire process.

It is important to \textbf{parameterize the score and LL models differently}. We parametrize the LL and the score using different models. Although there is a connection between them, i.e., the LL's gradient is the score according to the definition $s_t(\bx) = \nabla_{\bx} q_t(\bx)$, we obtain the score function beforehand in the fitting process using score matching or Score-PINN. Based on this, we can solve LL via the LL-ODE. Hence, the score's precision directly determines the LL's accuracy through the LL-ODE. It is not the case that there is a mutual regularization effect between score matching and the LL ODE. Consequently, we parameterize the score model and the LL model differently.

\subsection{Comparison between SM, SSM, and Score-PINN}\label{sec:comparison}
This subsection compares the pros and cons of SM, SSM, and Score-PINN, which are three distinct methods for solving the score that can be used to infer the LL and PDF and enable fast sampling.

Regarding speed, SM is the fastest, followed by SSM, while Score-PINN is the slowest. This is because SM involves only the inference of the score model itself, i.e., zero-order derivatives. SSM involves first-order derivatives, and Score-PINN involves second-order derivatives. With other hyperparameters being equal, especially under the premise of the same model structure and the same number of residual points sampled from the SDE trajectories, SM and SSM are equivalent and exhibit similar performance. Score-PINN, utilizing higher-order information, may achieve better results.

As for the comparison of the effectiveness between SM and SSM, mathematically, they are equivalent, and thus, their performance should be very close. However, in practice, due to SM utilizing the score of the conditional distribution for score matching, a weighting function $\lambda(t) = \sqrt{t}$ is needed to ensure numerical stability. The choice of this weighting function leads to the fact that the weights are small for smaller values of $t$, while for larger $t$, the weights are large, potentially leading to the slight difference between SM's and SSM's performance.

As previously mentioned, directly employing a full Score-PINN is excessively costly, and using the HTE \cite{song2021scorebased,hu2023hutchinson} to accelerate Score-PINN is generally preferred. This introduces a tradeoff for Score-PINN. According to the HTE error analysis in Hu et al. \cite{hu2023hutchinson}, HTE is relatively accurate in low dimensions, allowing Score-PINN to outperform SM and SSM, thanks to higher-order supervision. However, in high dimensions, the accuracy of HTE gradually diminishes, potentially causing Score-PINN to perform worse than SM and SSM. 

Regarding the applicability of the methods, SSM and Score-PINN are versatile, as the form of the SDE does not constrain their applicability. In contrast, SM, although cost-effective, requires the conditional probability of the SDE $p_{0t}(\bx | \bx_0)$, which is often challenging to achieve, thereby limiting its applicability.

Our computational experiments in Section \ref{sec:experiment} will confirm all these comparisons.

\subsection{Comparison with Direct Learning LL with PINN}\label{sec:ll_pinn}
We propose learning the score of SDE first and then solving the LL ODE based on the learned score to obtain the LL. This approach avoids numerical overflow issues associated with directly handling the PDF. A natural comparison is to examine the case of not learning the score and instead directly fitting the LL. However, directly learning the LL transforms the original linear FP equation into a nonlinear Hamilton-Jacobi-Bellman (HJB) equation, which cannot be solved using conventional mean square error loss and exhibits rapidly increasing errors with dimensionality \cite{wang20222}. 
Furthermore, we need to solve the HJB equation over the entire space $\mathbb{R}^d$, lacking boundary conditions, making it challenging to determine the characteristics of the solution at infinity. Consequently, vanilla PINN is prone to failure. However, score matching involves fitting the score solely on the SDE trajectories, eliminating the issue of boundary conditions. This makes fitting the score easier than directly fitting the LL. Once the score is obtained, acquiring the LL only requires solving time-dependent ODEs, significantly reducing the difficulty of obtaining the LL.
Our experiments in Section \ref{exp:ll_pinn} will demonstrate that the error in the score-based approach does not escalate with dimensionality, showcasing the robustness of learning the score compared to directly modeling the LL. 
Additionally, the numerical stability of the score is superior to that of the LL. For example, in the case of a Gaussian distribution, the score is a linear function whose scale is independent of dimension. At the same time, the LL is a polynomial with numerical growth linearly dependent on dimension. Under the Gaussian example, fitting it is also more straightforward due to the simplicity of the score. We will illustrate in our experiments in Section \ref{exp:ll_pinn} that the score-based model outperforms the vanilla PINN directly fitting the LL by a significant margin.

\subsection{Difference from Score-Based Generative Models}
Here, we compare our score-based SDE solver and score-based generative models \cite{song2021scorebased}. We solve the SDE and the FP equation via score. On the other hand, generative models generate data through the SDE without explicitly focusing on the PDF. Additionally, when we solve the SDE, we have known initial conditions and the form of the SDE, specifically the drift and diffusion coefficient. In generative models, the initial condition is the unknown data distribution. The generative model employs the SDE to transform the data distribution into Gaussian noise and then reverse the SDE. The reverse SDE serves as a pathway from noise to image, enabling image generation, and its drift coefficient is related to the score function. Hence, generative models conduct score matching for SDE reversal and data generation. Lastly, generative models use relatively simple SDEs for image generation, whereas we can solve arbitrarily complex SDEs. Overall, our score-based SDE solver provides a novel perspective for solving high-dimensional FP equations, while generative models create new samples via SDEs, which are totally different.

\section{Computational Experiments}\label{sec:experiment}
In this section, we conduct numerical experiments, thoroughly examining the score-based SDE solver's performance across various settings, including anisotropic Ornstein-Uhlenbeck (OU) processes, geometric Brownian motion, and Brownian motion with varying eigenspace. Considering the variance in SDEs, we explored variance-exploding SDEs and SDEs with finite variance. Depending on the probability distribution designed by the SDE solution, we examined a range of distributions, including Gaussian, Log-normal, Laplace, and Cauchy.

For experimental design, while we could enhance the complexity of the SDE by specifying more intricate drift and diffusion coefficients, traditional methods such as finite element and finite difference face the CoD when dealing with such general high-dimensional SDEs. The only viable alternative is Monte Carlo simulation, which entails expensive SDE discretization, extensive sampling, and inherent errors. Hence, relying on it for testing may not be reliable. Thus, we resort to analytically solvable SDEs.
In low dimensions, Monte Carlo sampling is commonly employed to construct histograms for testing, allowing for a visual comparison. However, such an approach becomes impractical in high dimensions.

We conduct all experiments on an NVIDIA A100 GPU with 80GB memory. We implement our algorithm by JAX \cite{jax2018github} due to its efficient automatic differentiation. The test metrics are relative $L_2$ and $L_\infty$ errors for the LL and PDF. All the experiments are computed five times with five independent random seeds, and we report the mean.

Regarding testing the PDF error in high dimensions, as previously mentioned, the numerical values of PDFs decay exponentially in high-dimensional spaces, rendering them extremely small. Consequently, our focus shifts to learning the LL. Nevertheless, we can still deduce the error in the PDF by examining the numerical values of the LL. Suppose that the exact LL on the test points is $\{q_t(\bx_i)\}_{i=1}^N$ and the predicted LL is $\{q_t(\bx_i;\theta)\}_{i=1}^N$. Without loss of generality, we assume that the exact LL is arranged in descending order, i.e., $q_t(\bx_1) > q_t(\bx_2) > \cdots > q_t(\bx_N)$. We normalize the exact values and the prediction by $q_t(\bx_1)$, whose values are denoted $\{\Tilde{q}_t(\bx_i)\}_{i=1}^N$ and $\{\Tilde{q}_t(\bx_i;\theta)\}_{i=1}^N$, where ${q}_t(\bx_i) = {q}_t(\bx_i) - {q}_t(\bx_1)$ and $\Tilde{q}_t(\bx_i;\theta) = {q}_t(\bx_i;\theta) - {q}_t(\bx_1)$. After normalization, they will be more numerically stable and can be directly used for PDF error calculation.

\subsection{Anisotropic Ornstein-Uhlenbeck Process}
\subsubsection{SDE Formulation}
We consider the Ornstein-Uhlenbeck (OU) process with anisotropic and correlated noises, and the initial condition $p_0(\bx)$ is unit Gaussian:
\begin{align}
d\bx = -\frac{1}{2}\bx dt + \bSigma^{\frac{1}{2}}d\bw_t,  \quad p_0(\bx) \sim \mathcal{N}(0, \bI).
\end{align}
The Brownian noise is correlated with the covariance matrix $\bSigma \in \mathbb{R}^{d \times d}$, which is constructed as follows.
\begin{itemize}
\item We generate random orthogonal matrix $\boldsymbol{Q}$ from QR decomposition, serving as the eigenspace of $\bSigma$.
\item $\bSigma$'s eigenvalues $\boldsymbol{\Gamma} = \operatorname{diag}(\lambda_1, \lambda_2, \cdots, \lambda_d)$ where $\lambda_{2i} \sim \text{Unif}[1, 1.1]$ and $\lambda_{2i+1} = 1 / \lambda_{2i}$.
\item Finally, $\bSigma = \boldsymbol{Q}^\mathrm{T}\boldsymbol{\Gamma}\boldsymbol{Q}$.
\end{itemize}
Thus, the SDE solution is anisotropic, which is a Gaussian $p_t(\bx) \sim \mathcal{N}\left(0, \exp(-t)\bI + (1 - \exp(-t))\bSigma\right) := \mathcal{N}(0, \bSigma_t)$ and the exact score function is $\bs_t(\bx) = \bSigma_t^{-1}\bx$ where $\bSigma_t = \exp(-t)\bI + (1 - \exp(-t))\bSigma$. The SDE has finite variance and gradually transforms the unit Gaussian to $\mathcal{N}(0,\bSigma)$ as $t \rightarrow \infty$. The score-matching and Score-PINN objectives can be derived from Algorithm \ref{algo:1}, as demonstrated in detail for the interested readers in Appendix \ref{appendix:exp1}.

\subsubsection{Hyperparameter Setting}
We test the SDE for $d=20, 50, 100$ cases and the terminal time $T=1$. 
The score model is a 4-layer fully connected network whose input and output dims are the same as the SDE dimensionality $d$, while the hidden dim is 512 for all cases. 
The LL model is also a 4-layer fully connected network whose input dimension is $d$, output dimension is 1, and hidden dimension is 512 for all cases.
The score and LL models are trained via Adam \cite{kingma2014adam} for 100K epochs, with an initial learning rate of 1e-3, which exponentially decays with a decay rate of 0.9 for every 10K epochs. We select 1K random residual points along the SDE trajectory at each Adam epoch and 10K fixed testing points for all methods in training score and LL. We adopt the following model structure to satisfy the initial condition with hard constraint and to avoid the boundary/initial loss \cite{lu2021physics} for the score and LL models
$
\bs_t(\bx; \theta) = \text{NN}(\bx, t; \theta) t - \bx,
q_t(\bx;\phi) = \text{NN}(\bx, t; \phi) t -\frac{d}{2}\log(2\pi) - \frac{1}{2}\bx^\mathrm{T}\bx,
$
where $\text{NN}(\bx, t; \theta)$ and $\text{NN}(\bx, t; \phi)$ are the fully connected neural network models, where we parameterize the score and LL differently based on Section \ref{sec:hidden_size}.

\subsubsection{Results}
\begin{table}[htbp]
\footnotesize
\centering
\begin{tabular}{|c|ccc|ccc|ccc|}
\hline
Dimension & \multicolumn{3}{c|}{20D} & \multicolumn{3}{c|}{50D} & \multicolumn{3}{c|}{100D} \\ \hline
Metric/Method & \multicolumn{1}{c|}{SM} & \multicolumn{1}{c|}{SSM} & Score-PINN & \multicolumn{1}{c|}{SM} & \multicolumn{1}{c|}{SSM} & Score-PINN & \multicolumn{1}{c|}{SM} & \multicolumn{1}{c|}{SSM} & Score-PINN \\ \hline
Speed & \multicolumn{1}{c|}{2704.81it/s} & \multicolumn{1}{c|}{471.25it/s} & 115.42it/s & \multicolumn{1}{c|}{2601.47it/s} & \multicolumn{1}{c|}{218.95it/s} & 59.81it/s & \multicolumn{1}{c|}{2415.43it/s} & \multicolumn{1}{c|}{121.02it/s} & 28.21it/s \\ \hline
LL $L_2$ Error & \multicolumn{1}{c|}{5.475E-3} & \multicolumn{1}{c|}{5.280E-3} & \textbf{2.379E-3} & \multicolumn{1}{c|}{6.394E-3} & \multicolumn{1}{c|}{5.215E-3} & \textbf{3.693E-3} & \multicolumn{1}{c|}{6.089E-3} & \multicolumn{1}{c|}{5.801E-3} & \textbf{5.289E-3} \\ \hline
LL $L_\infty$ Error & \multicolumn{1}{c|}{3.953E-2} & \multicolumn{1}{c|}{3.710E-2} & \textbf{2.291E-2} & \multicolumn{1}{c|}{3.721E-2} & \multicolumn{1}{c|}{3.281E-2} & \textbf{2.944E-2} & \multicolumn{1}{c|}{4.180E-2} & \multicolumn{1}{c|}{\textbf{3.167E-2}} & 3.235E-2 \\ \hline
PDF $L_2$ Error & \multicolumn{1}{c|}{1.270E-2} & \multicolumn{1}{c|}{1.611E-2} & \textbf{1.034E-2} & \multicolumn{1}{c|}{2.396E-2} & \multicolumn{1}{c|}{2.727E-2} & \textbf{2.350E-2} & \multicolumn{1}{c|}{4.022E-2} & \multicolumn{1}{c|}{\textbf{3.920E-2}} & 4.087E-2 \\ \hline
PDF $L_\infty$ Error & \multicolumn{1}{c|}{1.149E-2} & \multicolumn{1}{c|}{1.577E-02} & \textbf{1.117E-2} & \multicolumn{1}{c|}{2.181E-2} & \multicolumn{1}{c|}{2.447E-2} & \textbf{1.672E-2} & \multicolumn{1}{c|}{3.922E-2} & \multicolumn{1}{c|}{3.985E-2} & \textbf{3.839E-2} \\ \hline
\end{tabular}
\caption{Results for anisotropic OU process in 20D, 50D, and 100D.}
\label{tab:exp1}
\end{table}
The results are shown in Table \ref{tab:exp1}, where we report each algorithm's speed for score training (iteration per second) and their errors (relative $L_2$ and $L_\infty$ errors for LL and PDF). We only care about the speed of score training since the LL inference of all methods is the same, i.e., these methods only differ in how they obtain the score function.
In terms of speed, SM is faster than SSM, and SSM is faster than Score-PINN. As SM does not involve the derivatives of the score model, its speed remains nearly constant with increasing dimensionality. On the other hand, both SSM and Score-PINN calculate derivatives of the score model, resulting in a linear increase in time consumption with dimensionality. 
Regarding performance, since both SM and SSM are extensions of the score-matching objective, they are nearly equivalent and exhibit similar behavior.
Moreover, we observe that at relatively low dimensions (20D and 50D), Score-PINN significantly outperforms SM and SSM in all four metrics. However, as the dimensionality increases, since the accuracy of HTE in Score-PINN diminishes, the Score-PINN's advantage gradually diminishes as well, and the three methods, SM, SSM, and Score-PINN, perform similarly.
We also note that the performance of SM, SSM, and Score-PINN remains stable across dimensions: the LL $L_2$ error is roughly 5E-3, and other errors are 2E-2 to 4E-2.
These results demonstrate the capability of score-based SDE solvers to overcome the curse of dimensionality, whose computational cost only grows linearly and performance is stable across different dimensions. We also verify the analysis in Section \ref{sec:comparison} on method comparison. In theory, although our score-based SDE solvers can be applied to higher dimensions thanks to the algorithmic efficiency and stable performance, testing on PDF becomes gradually intractable due to its exponentially decaying values. Hence, in very high dimensions, testing PDF values will be very sensitive and, thus, less meaningful.

\subsection{Anisotropic Brownian Motion with Varying Eigenspace}
\subsubsection{SDE Formulation}
We consider the anisotropic Brownian motion and unit Gaussian as the initial condition:
\begin{align}
d\bx = (\bB + t \bI) d\bw_t, \quad p_0(\bx) \sim \mathcal{N}(0, \bI).
\end{align}
The exact solution is also a Gaussian $p_t(\bx) \sim \mathcal{N}(0, \boldsymbol{\Sigma}_t)$ and the exact score function is $\bs_t(\bx) = -\bSigma_t^{-1}\bx$, where
\begin{align}
\boldsymbol{\Sigma}_t &= \bI + \int_0^t \left[ (\bB +s \bI) (\bB +s \bI)^\mathrm{T}\right]ds  = \left(1 + \frac{t^3}{3}\right)\bI + t\bB\bB^\mathrm{T} + \frac{t^2}{2}(\bB + \bB^\mathrm{T}).
\end{align}
Here, we generate $\bB = \boldsymbol{Q}\boldsymbol{\Gamma}$ where $\boldsymbol{Q}$ is an orthogonal matrix and $\boldsymbol{\Gamma}$ is a diagonal matrix generated in the same way used in the previous example of anisotropic OU process. 
Concretely, the solution to this SDE is an anisotropic Gaussian, and its covariance matrix's eigenspace evolves as long as $\bB$ is not orthogonal and not symmetric. This complexity prevents us from simplifying the original SDE to an isotropic or lower-dimensional problem through elementary transformations or coordinate changes, highlighting the difficulty of our chosen SDE. We demonstrate the score-matching and score-PINN objectives in detail for the interested readers in Appendix \ref{appendix:varying_eigenspace}.

\subsubsection{Hyperparameter Setting}
We adopt the same hyperparameter setting as in the previous experiment. Notably, the vanilla score matching method matches the score function model with the conditional score $\nabla_{\bx}p_{0t}(\bx | \bx_0)$ via minimizing the mean square error between them. However, the conditional score of this SDE, namely $\nabla_{\bx}p_{0t}(\bx | \bx_0) = (\bSigma_t - \bI)^{-1}(\bx - \bx_0)$, cannot be easily computable due to the inverse $(\bSigma_t - \bI)^{-1} \in \mathbb{R}^{d \times d}$ for different sampled $t$, and it thus cannot enable efficient training. Hence, vanilla score matching is intractable in this case, and we will only report SSM and Score-PINN performances.

\subsubsection{Results}\label{sec:4.2.4}
\begin{table}[htbp]
\centering
\begin{tabular}{|c|cc|cc|cc|}
\hline
Dimension & \multicolumn{2}{c|}{20D} & \multicolumn{2}{c|}{50D} & \multicolumn{2}{c|}{100D} \\ \hline
Metric/Method & \multicolumn{1}{c|}{SSM} & Score-PINN & \multicolumn{1}{c|}{SSM} & Score-PINN & \multicolumn{1}{c|}{SSM} & Score-PINN \\ \hline
Speed & \multicolumn{1}{c|}{480.95it/s} & 116.78it/s & \multicolumn{1}{c|}{218.34it/s} & 49.65it/s & \multicolumn{1}{c|}{117.68it/s} & 27.41it/s \\ \hline
LL $L_2$ Error & \multicolumn{1}{c|}{1.923E-2} & \textbf{2.355E-3} & \multicolumn{1}{c|}{6.536E-3} & \textbf{5.471E-3} & \multicolumn{1}{c|}{1.109E-2} & \textbf{8.747E-3} \\ \hline
LL $L_\infty$ Error & \multicolumn{1}{c|}{6.929E-2} & \textbf{7.372E-3} & \multicolumn{1}{c|}{\textbf{1.962E-2}} & 2.013E-2 & \multicolumn{1}{c|}{3.645E-2
} & \textbf{3.085E-2} \\ \hline
PDF $L_2$ Error & \multicolumn{1}{c|}{2.561E-2} & \textbf{3.754E-3} & \multicolumn{1}{c|}{1.590E-2} & \textbf{1.380E-2} & \multicolumn{1}{c|}{\textbf{1.751E-2}} & 2.071E-2 \\ \hline
PDF $L_\infty$ Error & \multicolumn{1}{c|}{2.503E-2} & \textbf{3.229E-3} & \multicolumn{1}{c|}{1.599E-2} & \textbf{1.319E-2} & \multicolumn{1}{c|}{\textbf{1.460E-2}} & 1.999E-2 \\ \hline
\end{tabular}
\caption{Results for anisotropic Brownian motion with varying eigenspace in 20D, 50D, and 100D.}
\label{tab:exp2}
\end{table}
The results are presented in Table \ref{tab:exp2}, where we report each algorithm's speed for score training (iterations per second) and their errors (relative $L_2$ and $L_\infty$ errors for LL and PDF). Regarding speed, SSM outperforms Score-PINN, and the time consumption of both methods increases linearly with dimensionality. Regarding performance, we observe that at relatively low dimensions (20D), Score-PINN significantly outperforms SSM, surpassing SSM in all four metrics. However, as the dimensionality increases, the accuracy of HTE in Score-PINN diminishes, and Score-PINN's advantage gradually diminishes as well, converging towards the performance of SSM. We also note that the performance of SSM and Score-PINN remains stable across dimensions, generally falling within the range of 1E-2 to 1E-3, demonstrating the capability of score-based SDE solvers to overcome the CoD  whose computational cost only grows linearly, and performance is stable across different dimensions. We elaborate on this analysis in Section \ref{sec:comparison} on method comparison.

\subsubsection{Additional Study: Comparison with Direct Learning LL with Vanilla PINN}\label{exp:ll_pinn}
This additional study will validate our analysis in Section \ref{sec:ll_pinn}, i.e., Score-PINN outperforms vanilla PINN fitting the LL directly (Direct LL-PINN). We adopt the same model structure, training points, optimizer, scheduler, etc., for Direct LL-PINN and Score-PINN for a fair comparison. Furthermore, we apply adversarial training/$L_\infty$ loss to Direct LL-PINN according to Wang et al. \cite{wang20222}. We optimize for the training points using adversarial training for five epochs with a step size of 0.2 using an Adam optimizer with a 1e-3 learning rate.

The results are shown in Table \ref{tab:direct_ll_pinn}. Our Score-PINN outperforms Direct LL-PINN in all dimensions and metrics, with the latter's error escalating rapidly with increasing dimensions. In fact, Wang et al.'s \cite{wang20222} original findings also demonstrated that the (Direct LL-)PINN's error on the HJB equation increases with dimensionality. The HJB equation is obtained from applying the logarithm transform to the FP equation to directly fit the LL with Direct LL-PINN in Wang et al. \cite{wang20222}. In contrast, our study showcases the performance stability of Score-PINN across various high dimensions, significantly outperforming the Direct-LL-PINN in Wang et al. \cite{wang20222}. 
In addition to the poor performance of vanilla PINN, vanilla PINN is also about three times slower than Score-PINN. This is because vanilla PINN requires adversarial training. Originally, the Fokker-Planck equation and Score-PDE are of second-order, but adversarial training in vanilla PINN requires an additional order computation of derivative, which is third-order derivatives in total. In contrast, Score-PINN only requires the computation of second-order derivatives.

\begin{table}[htbp]
\small
\centering
\begin{tabular}{|c|cc|cc|cc|}
\hline
Dimension & \multicolumn{2}{c|}{20D} & \multicolumn{2}{c|}{50D} & \multicolumn{2}{c|}{100D} \\ \hline
Metric/Method & \multicolumn{1}{c|}{Vanilla-PINN} & Score-PINN & \multicolumn{1}{c|}{Vanilla-PINN} & Score-PINN & \multicolumn{1}{c|}{Vanilla-PINN} & Score-PINN \\ \hline
LL $L_2$ Error & \multicolumn{1}{c|}{4.917E-2} & \textbf{2.355E-3} & \multicolumn{1}{c|}{1.342E-1} & \textbf{5.471E-3} & \multicolumn{1}{c|}{4.769E-1} & \textbf{8.747E-3} \\ \hline
LL $L_\infty$ Error & \multicolumn{1}{c|}{8.715E-2} & \textbf{7.372E-3} & \multicolumn{1}{c|}{3.510E-1} & \textbf{2.013E-2} & \multicolumn{1}{c|}{8.946E-1
} & \textbf{3.085E-2} \\ \hline
PDF $L_2$ Error & \multicolumn{1}{c|}{6.791E-2} & \textbf{3.754E-3} & \multicolumn{1}{c|}{3.915E-1} & \textbf{1.380E-2} & \multicolumn{1}{c|}{6.251E-1} & \textbf{2.071E-2} \\ \hline
PDF $L_\infty$ Error & \multicolumn{1}{c|}{7.825E-2} & \textbf{3.229E-3} & \multicolumn{1}{c|}{3.870E-1} & \textbf{1.319E-2} & \multicolumn{1}{c|}{5.730E-1} & \textbf{1.999E-2} \\ \hline
\end{tabular}
\caption{Results for anisotropic Brownian motion with varying eigenspace in 20D, 50D, and 100D, comparing vanilla PINN (Direct LL-PINN) with Score-PINN.}
\label{tab:direct_ll_pinn}
\end{table}

\subsection{Anisotropic Ornstein-Uhlenbeck Process with Non-Gaussian Solution: Cauchy and Laplace Distributions}
\subsubsection{SDE Formulation}
We consider the anisotropic OU process with non-Gaussian distributions (Cauchy and  Laplace distributions) as the initial condition:
\begin{align}
\text{Cauchy Case:}\quad d\bx &= -\frac{1}{2}\bx dt + \bSigma^{\frac{1}{2}} d\bw_t, \quad p_0(\bx) = \frac{\Gamma\left(\frac{1 + d}{2}\right)}{\Gamma\left(\frac{1}{2}\right)\pi^{\frac{d}{2}}\left[1 + \Vert\bx\Vert^2\right]^{\frac{d+1}{2}}},\\
\text{Laplace Case:}\quad d\bx &= -\frac{1}{2}\bx dt + \bSigma^{\frac{1}{2}} d\bw_t, \quad p_0(\bx) = \frac{1}{2^d}\prod_{i=1}^d\exp\left(-|\bx_i|\right),
\end{align}
where $\Gamma(\cdot)$ denotes the Gamma function, the covariance matrix $\bSigma$ of the Brownian motion is generated as in the first experiment, and $p_0(\bx)$ is either the Cauchy distribution or the  Laplace PDF. The exact solution is the sum of an anisotropic Gaussian and the Cauchy/ Laplace distribution. In this context, we chose the initial heavy-tailed Cauchy/ Laplace distribution, significantly different from the Gaussian distribution. This choice allows us to test whether our score matching and Score-PINN can fit a distribution entirely distinct from Gaussian, which, in this example, is the sum of the Cauchy/Laplace and anisotropic Gaussian distributions. Moreover, the PDF of the Cauchy distribution is inseparable, and its various dimensional variables are not independent. The  Laplace PDF is also non-smooth and non-radial as Gaussian. Therefore, the effective dimensionality of this SDE is high due to the complicated initial distribution, and it cannot be simplified into a lower-dimensional problem, posing a significant challenge. We demonstrate the score-matching and Score-PINN objectives in detail for the interested readers in Appendix \ref{appendix:cauchy}.

\subsubsection{Hyperparameter Setting}
We test the SDE in 10D, 20D, and 30D at terminal time $T = 1$. We do not extend the testing to higher dimensions due to the limited precision in obtaining the reference solution through Monte Carlo simulation, as indicated in Appendix \ref{appendix:mc} where errors become substantial in 50 dimensions for Monte Carlo simulations. The score model is a 4-layer fully connected network whose input and output dimensions are the same as the SDE dimensionality $d$, while the hidden dimension is 128 for all cases. 
The LL model is also a 4-layer fully connected network whose input dimension is $d$, output dimension is 1, and hidden dimension is also 128 for all cases.
The score and LL models are trained via Adam \cite{kingma2014adam} for 100K epochs, with an initial learning rate of 1e-3, which exponentially decays with a decay rate of 0.9 for every 10K epochs. We select 1K random residual points along the SDE trajectory at each Adam epoch and 10K fixed testing points for all methods in training score and LL. The reference LL and PDF are generated via Monte Carlo simulation with $10^7$ samples. 

For the Cauchy case, we encode the initial condition into the model structure to satisfy the initial condition with hard constraint and to avoid the boundary/initial loss \cite{lu2021physics} for the score and LL models
$
\bs_t(\bx; \theta) = \text{NN}(\bx, t; \theta) t -\frac{(d + 1)\bx}{1 + \Vert \bx \Vert^2},
q_t(\bx;\phi) = \text{NN}(\bx, t; \phi) t+\log\left(\Gamma\left(\frac{1 + d}{2}\right)\right) - \log\left(\Gamma\left(\frac{1}{2}\right)\right) - \frac{d}{2}\log\pi - \frac{d+1}{2}\log\left(1 + \Vert\bx\Vert^2\right),
$
where $\text{NN}(\bx, t; \theta)$ and $\text{NN}(\bx, t; \phi)$ are the fully connected neural network models, where we parameterize the score and LL differently based on Section \ref{sec:hidden_size}.

Since the Laplace PDF and LL are non-smooth, we do not adopt the initial-augmented model structure to satisfy the initial condition automatically. Instead, we use the vanilla neural networks and adopt the boundary/initial loss with 20 weight to enforce the initial condition, and we assign unity loss to the residual loss.

\subsubsection{Results}
The results are presented in Table \ref{tab:brownian_cauchy}. The experimental results here align well with the findings in the preceding sections. In terms of speed, SM surpasses SSM, and SSM outpaces Score-PINN. Regarding effectiveness, Score-PINN outperforms both SM and SSM. This demonstrates the capability of score-based SDE solvers in handling complex SDEs that are non-Gaussian and anisotropic.
\begin{table}[htbp]
\footnotesize
\centering
\begin{tabular}{|c|ccc|ccc|ccc|}
\hline
Dimension & \multicolumn{3}{c|}{10D Cauchy} & \multicolumn{3}{c|}{20D Cauchy} & \multicolumn{3}{c|}{30D Cauchy} \\ \hline
Metric/Method & \multicolumn{1}{c|}{SM} & \multicolumn{1}{c|}{SSM} & Score-PINN & \multicolumn{1}{c|}{SM} & \multicolumn{1}{c|}{SSM} & Score-PINN & \multicolumn{1}{c|}{SM} & \multicolumn{1}{c|}{SSM} & Score-PINN \\ \hline
Speed & \multicolumn{1}{c|}{708.82it/s} & \multicolumn{1}{c|}{798.73it/s} & 629.58it/s & \multicolumn{1}{c|}{698.25it/s} & \multicolumn{1}{c|}{403.19it/s} & 339.89it/s & \multicolumn{1}{c|}{690.75it/s} & \multicolumn{1}{c|}{226.50it/s} & 194.34it/s \\ \hline
LL $L_2$ Error & \multicolumn{1}{c|}{3.692E-3} & \multicolumn{1}{c|}{3.244E-3} & \textbf{1.983E-3} & \multicolumn{1}{c|}{3.574E-3} & \multicolumn{1}{c|}{3.061E-3} & \textbf{2.172E-3} & \multicolumn{1}{c|}{3.682E-3} & \multicolumn{1}{c|}{3.815E-3} & \textbf{2.423E-3} \\ \hline
LL $L_\infty$ Error & \multicolumn{1}{c|}{7.513E-3} & \multicolumn{1}{c|}{9.281E-3} & \textbf{4.820E-3} & \multicolumn{1}{c|}{7.794E-3} & \multicolumn{1}{c|}{1.187E-2} & \textbf{5.976E-3} & \multicolumn{1}{c|}{7.816E-3} & \multicolumn{1}{c|}{7.680E-3} & \textbf{5.042E-3} \\ \hline
PDF $L_2$ Error & \multicolumn{1}{c|}{3.098E-2} & \multicolumn{1}{c|}{3.336E-2} & \textbf{2.913E-3} & \multicolumn{1}{c|}{3.675E-2} & \multicolumn{1}{c|}{3.259E-2} & \textbf{3.817E-3} & \multicolumn{1}{c|}{2.629E-2} & \multicolumn{1}{c|}{2.645E-2} & \textbf{1.926E-3} \\ \hline
PDF $L_\infty$ Error & \multicolumn{1}{c|}{4.871E-2} & \multicolumn{1}{c|}{4.602E-2} & \textbf{1.872E-3} & \multicolumn{1}{c|}{5.677E-2} & \multicolumn{1}{c|}{3.206E-2} & \textbf{3.551E-3} & \multicolumn{1}{c|}{2.358E-2} & \multicolumn{1}{c|}{2.986E-2} & \textbf{2.088E-3} \\ \hline
\hline
Dimension & \multicolumn{3}{c|}{10D Laplace} & \multicolumn{3}{c|}{20D Laplace} & \multicolumn{3}{c|}{30D Laplace} \\ \hline
Metric/Method & \multicolumn{1}{c|}{SM} & \multicolumn{1}{c|}{SSM} & Score-PINN & \multicolumn{1}{c|}{SM} & \multicolumn{1}{c|}{SSM} & Score-PINN & \multicolumn{1}{c|}{SM} & \multicolumn{1}{c|}{SSM} & Score-PINN \\ \hline
Speed & \multicolumn{1}{c|}{708.82it/s} & \multicolumn{1}{c|}{798.73it/s} & 629.58it/s & \multicolumn{1}{c|}{698.25it/s} & \multicolumn{1}{c|}{403.19it/s} & 339.89it/s & \multicolumn{1}{c|}{690.75it/s} & \multicolumn{1}{c|}{226.50it/s} & 194.34it/s \\ \hline
LL $L_2$ Error & \multicolumn{1}{c|}{5.424E-3} & \multicolumn{1}{c|}{4.509E-3} & \textbf{2.456E-3} & \multicolumn{1}{c|}{5.107E-3} & \multicolumn{1}{c|}{5.614E-3} & \textbf{3.307E-3} & \multicolumn{1}{c|}{4.213E-3} & \multicolumn{1}{c|}{6.974E-3} & \textbf{3.008E-3} \\ \hline
LL $L_\infty$ Error & \multicolumn{1}{c|}{6,757E-3} & \multicolumn{1}{c|}{8.595E-3} & \textbf{5.225E-3} & \multicolumn{1}{c|}{7.348E-3} & \multicolumn{1}{c|}{9.012E-3} & \textbf{4.510E-3} & \multicolumn{1}{c|}{8.699E-3} & \multicolumn{1}{c|}{1.460E-2} & \textbf{5.135E-3} \\ \hline
PDF $L_2$ Error & \multicolumn{1}{c|}{1.036E-2} & \multicolumn{1}{c|}{1.176E-2} & \textbf{5.974E-3} & \multicolumn{1}{c|}{1.390E-2} & \multicolumn{1}{c|}{9.403E-3} & \textbf{6.540E-3} & \multicolumn{1}{c|}{1.318E-2} & \multicolumn{1}{c|}{9.438E-3} & \textbf{6.367E-3} \\ \hline
PDF $L_\infty$ Error & \multicolumn{1}{c|}{2.057E-2} & \multicolumn{1}{c|}{1.855E-2} & \textbf{6.475E-3} & \multicolumn{1}{c|}{1.869E-2} & \multicolumn{1}{c|}{2.406E-2} & \textbf{4.535E-3} & \multicolumn{1}{c|}{1.682E-2} & \multicolumn{1}{c|}{1.705E-2} & \textbf{5.921E-3} \\ \hline
\end{tabular}
\caption{Results for anisotropic OU process with non-Gaussian solution in 10D, 20D, and 30D.}
\label{tab:brownian_cauchy}
\end{table}

\subsection{Geometric Brownian Motion with Anisotropic Log-Normal Distribution}
\subsubsection{SDE Formulation}
We consider the geometric Brownian motion whose initial condition $p_0(\bx)$ is an anisotropic Log-normal distribution:
\begin{align}
d\bx = \frac{1}{2}e^{-t}\bx dt + e^{-t/2}\bx d\bw_t, \quad p_0(\bx) = \frac{1}{(2\pi)^{d/2}\operatorname{det}(\bSigma)^{1/2}}\left(\prod_{i=1}^d \frac{1}{\bx_i}\right)\exp\left(-\frac{(\log\bx)^\mathrm{T}\bSigma^{-1}(\log\bx)}{2}\right).
\end{align}
The covariance matrix $\bSigma$ of the initial distribution is generated from the same procedure as the first experiment.
The SDE solution is an anisotropic Log-normal distribution with mean zero and variance $\bSigma_t = \bSigma + (1 - \exp(-t))\bI$. 
This problem is even more challenging compared to the previous ones. Previously, we attempted initial distributions with long tails, such as Cauchy and Laplace distributions, but the OU process would inject short-tail Gaussian noise. Here, both the initial distribution and the SDE-injected noise are long-tailed log-normal distributions, adding additional difficulty. Moreover, this dual long-tail nature significantly reduces the accuracy of Monte Carlo simulations; see Appendix \ref{appendix:mc2}.
We demonstrate the score-matching and Score-PINN objectives in detail  for interested readers in
Appendix \ref{appendix:gbm}.

\subsubsection{Hyperparameter Setting}
We test the SDE in 10D, 20D, and 30D at terminal time $T = 0.3/0.1$. We do not extend the testing to higher dimensions and longer time problems due to the ill-posedness of Log-normal with a long tail at $\bx \rightarrow \infty$, as indicated in Appendix \ref{appendix:mc2}. The score model is a 4-layer fully connected network whose input and output dims are the same as the SDE dimensionality $d$, while the hidden dimension is 128 for all cases. 
The LL model is also a 4-layer fully connected network whose input dimension is $d$, output dimension is 1, and hidden dimension is also 128 for all cases.
The score and LL models are trained via Adam \cite{kingma2014adam} for 100K epochs, with an initial learning rate of 1e-3, which exponentially decays with a decay rate of 0.9 for every 10K epochs. We select 1K random residual points along the SDE trajectory at each Adam epoch and 10K fixed testing points for all methods in training score and LL. We adopt the following model structure to satisfy the initial condition with hard constraint and to avoid the boundary/initial loss \cite{lu2021physics} for the score and LL models
$
\bs_t(\bx; \theta) = \text{NN}(\bx, t; \theta) t - \nabla_{\bx} \log p_0(\bx),
q_t(\bx;\phi) = \text{NN}(\bx, t; \phi) t - \log p_0(\bx),
$
where $\text{NN}(\bx, t; \theta)$ and $\text{NN}(\bx, t; \phi)$ are the fully connected neural network models, where we parameterize the score and LL differently based on Section \ref{sec:hidden_size}. Here, $p_0(\bx)$ is the initial distribution, which is an anisotropic Log-normal in this case.

\subsubsection{Results}
The results are presented in Table \ref{tab:gbm_log_normal}.
We tested for $T=0.3$ or $0.1$, where longer time poses greater difficulty. Score-PINN outperforms in all problem settings due to its utilization of high-order information, and its performance remains relatively stable across different dimensions. Notably, in the challenging setting of $T=0.3$, SM and SSM exhibit imprecise fitting of the score function, leading to divergence, while Score-PINN maintains reasonable accuracy. This reflects the significant challenge the Log-normal distribution poses as an ill-posed heavy-tailed distribution with support in positive real numbers for our score-based SDE solver. In fact, even using Monte Carlo simulation to solve problems related to geometric Brownian motion and Log-normal leads to substantial errors, see Appendix \ref{appendix:mc2}. In summary, this problem is highly challenging, where Score-PINN outperforms SM and SSM, and Monte Carlo simulation fails, demonstrating the advantage of a score-based SDE solver over traditional Monte Carlo methods that suffer from huge error due to the long tail of Log-normal distribution and geometric Brownian motion. The prediction of longer time intervals can be achieved via temporal domain decomposition and sequential time-marching strategies \cite{penwarden2023unified}.

\begin{table}[htbp]
\footnotesize
\centering
\begin{tabular}{|c|ccc|ccc|ccc|}
\hline
Dimension & \multicolumn{3}{c|}{10D Log-Normal $T=0.3$} & \multicolumn{3}{c|}{20D Log-Normal $T=0.3$} & \multicolumn{3}{c|}{30D Log-Normal $T=0.3$} \\ \hline
Metric/Method & \multicolumn{1}{c|}{SM} & \multicolumn{1}{c|}{SSM} & Score-PINN & \multicolumn{1}{c|}{SM} & \multicolumn{1}{c|}{SSM} & Score-PINN & \multicolumn{1}{c|}{SM} & \multicolumn{1}{c|}{SSM} & Score-PINN \\ \hline
Speed & \multicolumn{1}{c|}{723.81it/s} & \multicolumn{1}{c|}{800.40it/s} & 619.46it/s & \multicolumn{1}{c|}{721.04it/s} & \multicolumn{1}{c|}{400.19it/s} & 309.20it/s & \multicolumn{1}{c|}{700.23it/s} & \multicolumn{1}{c|}{199.81it/s} & 159.01it/s \\ \hline
LL $L_2$ Error & \multicolumn{1}{c|}{Diverge} & \multicolumn{1}{c|}{Diverge} & \textbf{4.720E-3} & \multicolumn{1}{c|}{Diverge} & \multicolumn{1}{c|}{Diverge} & \textbf{7.824E-3} & \multicolumn{1}{c|}{Diverge} & \multicolumn{1}{c|}{Diverge} & \textbf{6.793E-3} \\ \hline
LL $L_\infty$ Error & \multicolumn{1}{c|}{Diverge} & \multicolumn{1}{c|}{Diverge} & \textbf{3.106E-2} & \multicolumn{1}{c|}{Diverge} & \multicolumn{1}{c|}{Diverge} & \textbf{9.260E-2} & \multicolumn{1}{c|}{Diverge} & \multicolumn{1}{c|}{Diverge} & \textbf{8.020E-2} \\ \hline
PDF $L_2$ Error & \multicolumn{1}{c|}{Diverge} & \multicolumn{1}{c|}{Diverge} & \textbf{3.743E-2} & \multicolumn{1}{c|}{Diverge} & \multicolumn{1}{c|}{Diverge} & \textbf{4.022E-2} & \multicolumn{1}{c|}{Diverge} & \multicolumn{1}{c|}{Diverge} & \textbf{3.625E-2} \\ \hline
PDF $L_\infty$ Error & \multicolumn{1}{c|}{Diverge} & \multicolumn{1}{c|}{Diverge} & \textbf{3.972E-2} & \multicolumn{1}{c|}{Diverge} & \multicolumn{1}{c|}{Diverge} & \textbf{3.872E-2} & \multicolumn{1}{c|}{Diverge} & \multicolumn{1}{c|}{Diverge} & \textbf{3.509E-2} \\ \hline
\hline
Dimension & \multicolumn{3}{c|}{10D Log-Normal $T=0.1$} & \multicolumn{3}{c|}{20D Log-Normal $T=0.1$} & \multicolumn{3}{c|}{30D Log-Normal $T=0.1$} \\ \hline
Metric/Method & \multicolumn{1}{c|}{SM} & \multicolumn{1}{c|}{SSM} & Score-PINN & \multicolumn{1}{c|}{SM} & \multicolumn{1}{c|}{SSM} & Score-PINN & \multicolumn{1}{c|}{SM} & \multicolumn{1}{c|}{SSM} & Score-PINN \\ \hline
Speed & \multicolumn{1}{c|}{723.81it/s} & \multicolumn{1}{c|}{800.40it/s} & 619.46it/s & \multicolumn{1}{c|}{721.04it/s} & \multicolumn{1}{c|}{400.19it/s} & 309.20it/s & \multicolumn{1}{c|}{700.23it/s} & \multicolumn{1}{c|}{199.81it/s} & 159.01it/s \\ \hline
LL $L_2$ Error & \multicolumn{1}{c|}{3.391E-3} & \multicolumn{1}{c|}{3.025E-3} & \textbf{1.893E-3} & \multicolumn{1}{c|}{3.289E-3} & \multicolumn{1}{c|}{3.634E-3} & \textbf{1.923E-3} & \multicolumn{1}{c|}{3.826E-3} & \multicolumn{1}{c|}{3.232E-3} & \textbf{1.801E-3} \\ \hline
LL $L_\infty$ Error & \multicolumn{1}{c|}{6.047E-2} & \multicolumn{1}{c|}{5.980E-2} & \textbf{3.235E-2} & \multicolumn{1}{c|}{6.255E-2} & \multicolumn{1}{c|}{6.450E-2} & \textbf{2.058E-2} & \multicolumn{1}{c|}{5.006E-2} & \multicolumn{1}{c|}{4.683E-2} & \textbf{2.814E-2} \\ \hline
PDF $L_2$ Error & \multicolumn{1}{c|}{1.972E-2} & \multicolumn{1}{c|}{2.003E-2} & \textbf{8.201E-3} & \multicolumn{1}{c|}{2.261E-2} & \multicolumn{1}{c|}{2.311E-2} & \textbf{7.201E-3} & \multicolumn{1}{c|}{1.937E-2} & \multicolumn{1}{c|}{1.825E-2} & \textbf{5.560E-3} \\ \hline
PDF $L_\infty$ Error & \multicolumn{1}{c|}{1.825E-2} & \multicolumn{1}{c|}{1.978E-2} & \textbf{5.768E-3} & \multicolumn{1}{c|}{2.183E-2} & \multicolumn{1}{c|}{2.012E-2} & \textbf{5.034E-3} & \multicolumn{1}{c|}{1.773E-2} & \multicolumn{1}{c|}{1.762E-2} & \textbf{4.788E-3} \\ \hline
\end{tabular}
\caption{Results for geometric Brownian motion with Log-normal distribution and terminal time $T=0.3$ or $T=0.1$.}
\label{tab:gbm_log_normal}
\end{table}

\section{Summary}
In conclusion, our study addresses the persistent challenge of the curse-of-dimensionality (CoD) in high-dimensional Stochastic Differential Equations (SDEs) and their corresponding Fokker-Planck (FP) equations. By leveraging a score-based SDE solver to fit the score function, we present a novel approach that allows for the accurate inference of Logarithm-Likelihood (LL), Probability Density Function (PDF), and fast SDE simulation. Three fitting methods, Score Matching (SM), Sliced Score Matching (SSM), and Score-PINN, are introduced to learn the score function using physics-informed machine learning and thoroughly compared, considering computational complexity, accuracy, and generality. After approximation of the score function, fast PDF, LL, and SDE simulation inferences are enabled.
Our evaluations demonstrate the stability and performance of the proposed score-based SDE solver across diverse SDEs, demonstrating its applicability in addressing the CoD. Notably, our methodology outperforms traditional methods, maintaining stability in computational cost as dimensions increase. The Score-PINN method, in particular, stands out for its ability to leverage higher-order information, resulting in superior accuracy. On the other hand, SM and SSM, although their performances are slightly worse, exhibit much faster training than the Score-PINN due to the absence of costly high-order derivatives.
This research contributes a promising solution to the challenges associated with high-dimensional stochastic systems, opening avenues for further exploration and application across diverse scientific and engineering domains. 

\newpage

\appendix

\section{Experimental Details: SDE, Score PDE, and LL ODE Validation}

\subsection{Anisotropic Ornstein–Uhlenbeck (OU) Process}\label{appendix:exp1}
\subsubsection{Score Matching and Socre PINN Objective Losses}
For vanilla score matching, we match the score function model $\bs_t(\bx;\theta)$ with the conditional score $\nabla_{\bx}p_{0t}(\bx | \bx_0) = (\bSigma_t - \bI)^{-1}(\bx - \bx_0)$ using the objective function as in equation (\ref{eq:loss_sm}).
For sliced score matching, since it is agnostic to the SDE, the objective function in equation (\ref{eq:loss_ssm}) is adopted.
For score-PINN, the score PDE is given by
\begin{align}\label{eq:score-pde-exm1}
\partial_t \bs_t(\bx) &= \nabla_{\bx}\left[\frac{1}{2}\nabla_{\bx}\cdot\left(\bSigma\bs\right) + \frac{1}{2}\left\|\bSigma^{\frac{1}{2}}\bs\right\|^2 + \frac{1}{2}\langle\bx,\bs \rangle + \frac{d}{2}\right],\\
\bs_0(\bx) &= -\bx.
\end{align}
After obtaining the score function via either score matching or score-PINN, another LL-PINN can be used to infer LL from the LL ODE given by
\begin{align}\label{eq:ll-ode-exm1}
\partial_t q_t(\bx) &= \frac{1}{2}\nabla_{\bx}\cdot\left(\bSigma\bs\right) + \frac{1}{2}\left\|\bSigma^{\frac{1}{2}}\bs\right\|^2 + \frac{1}{2}\langle\bx,\bs \rangle + \frac{d}{2},\\
q_0(\bx) &= -\frac{d}{2}\log(2\pi) - \frac{1}{2}\bx^\mathrm{T}\bx,
\end{align}
where the exact LL gives the true solution $q_t(\bx) = -\frac{d}{2}\log(2\pi) - \frac{1}{2}\log\operatorname{det}(\bSigma_t) - \frac{1}{2}\bx^\mathrm{T}(\bSigma_t)^{-1}\bx$.

The mathematical validation of the Score-PDE and LL-PDE is presented as follows.
\subsubsection{Validation of the Score PDE and LL ODE}
\textbf{SDE and exact solution}. We consider the anisotropic OU process and unit Gaussian as the initial condition:
\begin{align}
d\bx = -\frac{1}{2}\bx dt + \bSigma^{\frac{1}{2}}d\bw_t, \quad p_0(\bx) \sim \mathcal{N}(0, \bI).
\end{align}
To solve the SDE analytically, we transform $\by = \exp(t / 2) \bx$, then $\by$ satisfy
\begin{align}
d\by = \frac{1}{2}\exp\left(\frac{t}{2}\right)\bx dt + \exp\left(\frac{t}{2}\right)d\bx = \exp\left(\frac{t}{2}\right)\bSigma^{\frac{1}{2}}d\bw_t, \quad p_0(\by) = \mathcal{N}(0, \bI).
\end{align}
Hence, the exact solution is $p_t(\bx) \sim \mathcal{N}\left(0, \exp(-t)\bI + (1 - \exp(-t))\bSigma\right) := \mathcal{N}(0, \bSigma_t)$ and the exact score function is $\bs_t(\bx) = \bSigma_t^{-1}\bx$ where $\bSigma_t = \exp(-t)\bI + (1 - \exp(-t))\bSigma$.
\\

\noindent\textbf{Validation of the score PDE \eqref{eq:score-pde-exm1}}. 
On the left-hand side, we use the fact that $\partial_t \bSigma_t^{-1} =\bSigma_t^{-1}(\partial_t\bSigma_t) \bSigma_t^{-1}$,
\begin{align}
\partial_t \bs_t(\bx) = -\partial_t \left(\bSigma_t^{-1} \bx\right) 
= \bSigma_t^{-1}\left(\partial_t \bSigma_t\right)\bSigma_t^{-1} \bx = \bSigma_t^{-1}\left(-\exp(-t)\bI + \exp(-t)\bSigma\right)\bSigma_t^{-1} \bx.
\end{align}
On the right-hand side,
\begin{align}
&\quad\nabla_{\bx}\left[\frac{1}{2}\nabla_{\bx}\cdot\left(\bSigma\bs\right) + \frac{1}{2}\left\|\bSigma^{\frac{1}{2}}\bs\right\|^2 + \frac{1}{2}\langle\bx,\bs \rangle + \frac{d}{2}\right] \\
&=\nabla_{\bx}\left[\frac{1}{2}\left\|\bSigma^{\frac{1}{2}}\bs\right\|^2 - \frac{1}{2}\bx^\mathrm{T}\bSigma_t^{-1}\bx\right] =\nabla_{\bx}\left[\frac{1}{2}\bx^\mathrm{T}\bSigma_t^{-1}\bSigma\bSigma_t^{-1}\bx - \frac{1}{2}\bx^\mathrm{T}\bSigma_t^{-1}\bx\right] = \bSigma_t^{-1}\bSigma\bSigma_t^{-1}\bx - \bSigma_t^{-1}\bx\\
&= \bSigma_t^{-1}\bSigma\bSigma_t^{-1}\bx - \bSigma_t^{-1}\bSigma_t\bSigma_t^{-1}\bx = \bSigma_t^{-1}\left(\bSigma - \bSigma_t\right)\bSigma_t^{-1}\bx = \bSigma_t^{-1}\left(-\exp(-t)\bI + \exp(-t)\bSigma\right)\bSigma_t^{-1} \bx.
\end{align}
\textbf{Validation of the LL ODE \eqref{eq:ll-ode-exm1}}. The exact LL  is 
given by
\begin{align}
q_t(\bx) &= -\frac{d}{2}\log(2\pi) - \frac{1}{2}\log\operatorname{det}(\bSigma_t) - \frac{1}{2}\bx^\mathrm{T}\bSigma_t^{-1}\bx.
\end{align}
On the right-hand side of \eqref{eq:ll-ode-exm1},
\begin{align}
&\quad\frac{1}{2}\nabla_{\bx} \cdot\left(\bSigma\bs\right) + \frac{1}{2}\Vert\bSigma^{\frac{1}{2}}\bs\Vert^2 + \frac{1}{2}\langle \bx, \bs\rangle +\frac{d}{2} \\
&=-\frac{1}{2}\operatorname{Tr}(\bSigma\bSigma_t^{-1}) + \frac{1}{2}\bx^\mathrm{T}\bSigma_t^{-1}\bSigma\bSigma_t^{-1}\bx - \frac{1}{2}\bx^\mathrm{T}\bSigma_t^{-1}\bx + \frac{d}{2}\\
&= -\frac{1}{2}\operatorname{Tr}(\bSigma\bSigma_t^{-1}) + \frac{1}{2}\bx^\mathrm{T}\bSigma_t^{-1}\left(\bSigma - \bSigma_t\right)\bSigma_t^{-1}\bx+ \frac{d}{2}.
\end{align}
On the left-hand side, we further use the fact that $\partial_t \log\operatorname{det}(\bSigma_t) = \operatorname{Tr}(\bSigma^{-1}\partial_t \bSigma_t)$,
\begin{align}
\partial_t \left(- \frac{1}{2}\bx^\mathrm{T}\bSigma_t^{-1}\bx\right) = \frac{1}{2}\bx^\mathrm{T}\bSigma^{-1}_t\left(-\exp(-t)\bI + \exp(-t)\bSigma\right)\bSigma^{-1}_t\bx = \frac{1}{2}\bx^\mathrm{T}\bSigma^{-1}_t\left(\bSigma - \bSigma_t\right)\bSigma^{-1}_t\bx
\end{align}
\begin{align}
\partial_t\left(- \frac{1}{2}\log\operatorname{det}(\bSigma_t)\right) = -\frac{1}{2}\operatorname{Tr}\left(\bSigma_t^{-1}\partial_t\bSigma_t\right) = -\frac{1}{2}\operatorname{Tr}\left(\bSigma_t^{-1}(\bSigma - \bSigma_t)\right) = -\frac{1}{2}\operatorname{Tr}\left(\bSigma_t^{-1}\bSigma\right) + \frac{d}{2}.
\end{align}

\subsection{Anisotropic Brownian with Varying Eigenspace}\label{appendix:varying_eigenspace}
\subsubsection{Score Matching, Score PDE, and LL ODE}
The vanilla score matching matches the score function model with the conditional score $\nabla_{\bx}p_{0t}(\bx | \bx_0)$ via minimizing the mean square error between them. However, the conditional score of this SDE, namely $\nabla_{\bx}p_{0t}(\bx | \bx_0) = (\bSigma_t - \bI)^{-1}(\bx - \bx_0)$, cannot be easily computable due to the inverse $(\bSigma_t - \bI)^{-1} \in \mathbb{R}^{d \times d}$ for different sampled $t$, and it thus cannot enable efficient training. Hence, vanilla score matching is intractable in this case.
On the other hand, since sliced score matching is agnostic to the SDE, the objective function in equation (\ref{eq:loss_ssm}) is adopted as usual.
For score-PINN, the score PDE is given by
\begin{align}\label{eq:score-pde-exm2}
\partial_t \bs_t(\bx) &= \nabla_{\bx}\left[\frac{1}{2}\nabla_{\bx}\cdot\left((\bB + t\bI)(\bB + t\bI)^\mathrm{T}\bs\right) + \frac{1}{2}\left\|(\bB + t\bI)^\mathrm{T}\bs\right\|^2\right],\\
\bs_0(\bx) &= -\bx,
\end{align}
where the solution to the score PDE is the exact score $\bs_t(\bx) = -\bSigma_t^{-1}\bx$.
The LL ODE is given by
\begin{align}\label{eq:ll-ode-exm2}
\partial_t q_t(\bx) &= \frac{1}{2}\nabla_{\bx} \cdot\left((\bB + t\bI)(\bB + t\bI)^\mathrm{T}\bs\right) + \frac{1}{2}\Vert(\bB + t\bI)^\mathrm{T}\bs\Vert^2,\\
q_0(\bx) &= -\frac{d}{2}\log(2\pi) - \frac{1}{2}\bx^\mathrm{T}\bx,
\end{align}
where the exact LL is the solution to the LL ODE, given by $q_t(\bx) = -\frac{d}{2}\log(2\pi) - \frac{1}{2}\log\operatorname{det}(\bSigma_t) - \frac{1}{2}\bx^\mathrm{T}\bSigma_t^{-1}\bx$. We theoretically validate the score PDE and the LL ODE as follows.

\subsubsection{Validation of the Score PDE and LL ODE}
\textbf{SDE and exact solution}. We consider the anisotropic Brownian motion and unit Gaussian as the initial condition:
\begin{align}
d\bx = (\bB + t \bI) d\bw_t, \quad p_0(\bx) \sim \mathcal{N}(0, \bI).
\end{align}
The exact solution is also a Gaussian $p_t(\bx) \sim \mathcal{N}(0, \boldsymbol{\Sigma}_t)$ and the exact score function is $\bs_t(\bx) = -\bSigma_t^{-1}\bx$, where
\begin{align}
\boldsymbol{\Sigma}_t &= \bI + \int_0^t \left[ (\bB +s \bI) (\bB +s \bI)^\mathrm{T}\right]ds  = \left(1 + \frac{t^3}{3}\right)\bI + t\bB\bB^\mathrm{T} + \frac{t^2}{2}(\bB + \bB^\mathrm{T}).
\end{align}
\textbf{Validation of the score PDE \eqref{eq:score-pde-exm2}}. 
On the left-hand side of \eqref{eq:score-pde-exm2},
\begin{align}
\partial_t \bs_t(\bx) = -\partial_t \left(\bSigma_t^{-1} \bx\right) 
= \bSigma_t^{-1}\left(\partial_t \bSigma_t\right)\bSigma_t^{-1} \bx = \bSigma_t^{-1}\left(t^2\bI + \bB\bB^\mathrm{T} + t(\bB + \bB^\mathrm{T})\right)\bSigma_t^{-1} \bx
\end{align}
On the right-hand side of \eqref{eq:score-pde-exm2},
\begin{align}
&\quad\nabla_{\bx}\left[\frac{1}{2}\nabla_{\bx}\cdot\left((\bB + t\bI)(\bB + t\bI)^\mathrm{T}\bs\right) + \frac{1}{2}\left\|(\bB + t\bI)^\mathrm{T}\bs\right\|^2\right] \\
&=\nabla_{\bx}\left[\frac{1}{2}\left\|(\bB + t\bI)^\mathrm{T}\bs\right\|^2\right] =\nabla_{\bx}\left[\frac{1}{2}\bx\bSigma^{-1}_t(\bB + t\bI)(\bB + t\bI)^\mathrm{T}\bSigma^{-1}_t\bx\right] = \bSigma^{-1}_t(\bB + t\bI)(\bB + t\bI)^\mathrm{T}\bSigma^{-1}_t\bx\\
&= \bSigma_t^{-1}\left(t^2\bI + \bB\bB^\mathrm{T} + t(\bB + \bB^\mathrm{T})\right)\bSigma_t^{-1} \bx
\end{align}
\textbf{Validation of the LL ODE}. The exact LL is  
given by
\begin{align}
q_t(\bx) = -\frac{d}{2}\log(2\pi) - \frac{1}{2}\log\operatorname{det}(\bSigma_t) - \frac{1}{2}\bx^\mathrm{T}\bSigma_t^{-1}\bx.
 \partial_t q_t(\bx) = \frac{1}{2}\nabla_{\bx} \cdot\left((\bB + t\bI)(\bB + t\bI)^\mathrm{T}\bs\right) + \frac{1}{2}\Vert(\bB + t\bI)^\mathrm{T}\bs\Vert^2.
\end{align}
On the right-hand side of \eqref{eq:ll-ode-exm2},
\begin{align}
&\quad\frac{1}{2}\nabla_{\bx}\cdot\left((\bB + t\bI)(\bB + t\bI)^\mathrm{T}\bs\right) + \frac{1}{2}\left\|(\bB + t\bI)^\mathrm{T}\bs\right\|^2 \\
&=-\frac{1}{2}\operatorname{Tr}\left((\bB + t\bI)(\bB + t\bI)^\mathrm{T}\bSigma_t^{-1}\right) + \frac{1}{2}\bx\bSigma^{-1}_t(\bB + t\bI)(\bB + t\bI)^\mathrm{T}\bSigma^{-1}_t\bx
\end{align}
On the left-hand side of \eqref{eq:ll-ode-exm2},
\begin{align}
\partial_t \left(- \frac{1}{2}\bx^\mathrm{T}\bSigma_t^{-1}\bx\right) &= \frac{1}{2}\bx\bSigma^{-1}_t(\bB + t\bI)(\bB + t\bI)^\mathrm{T}\bSigma^{-1}_t\bx\\
\partial_t\left(- \frac{1}{2}\log\operatorname{det}(\bSigma_t)\right) &= -\frac{1}{2}\operatorname{Tr}\left(\bSigma_t^{-1}\partial_t\bSigma_t\right) = -\frac{1}{2}\operatorname{Tr}\left((\bB + t\bI)(\bB + t\bI)^\mathrm{T}\bSigma_t^{-1}\right) .
\end{align}

\subsection{Anisotropic Ornstein-Uhlenbeck Process with Non-Gaussian Solution: Cauchy and Laplace Distributions}\label{appendix:cauchy}
This SDE is identical to our first experiment except for the initial condition/distribution difference. Hence, their score matching, score PDE, and LL ODE are similar except for the initial.
For vanilla score matching, we match the score function model with the conditional score $\nabla_{\bx}p_{0t}(\bx_t | \bx_0) = \left((1 - \exp(-t))(\bSigma - I)\right)^{-1}(\bx - \bx_0)$ using the objective function as in equation (\ref{eq:loss_sm}).
Since it is agnostic to the SDE, the objective function in equation (\ref{eq:loss_ssm}) is adopted for sliced score matching.
For score-PINN, the score PDE is given by
\begin{align}
\partial_t \bs_t(\bx) &= \nabla_{\bx}\left[\frac{1}{2}\nabla_{\bx}\cdot\left(\bSigma\bs\right) + \frac{1}{2}\left\|\bSigma^{\frac{1}{2}}\bs\right\|^2 + \frac{1}{2}\langle\bx,\bs \rangle + \frac{d}{2}\right],\\
\bs_0(\bx) &= -\frac{(d + 1)\bx}{1 + \Vert \bx \Vert^2} \quad \text{(Cauchy)}; \quad \bs_0(\bx) = -\log(2) - \sum_{i=1}^d\log(|\bx_i|)\quad \text{(Laplace)},
\end{align}
where the exact score is not analytically computable, requiring Monte Carlo simulation.
After obtaining the score function via either score matching or score-PINN, another LL-PINN can be used to infer LL from the LL ODE:
\begin{align}
\partial_t q_t(\bx) &= \frac{1}{2}\nabla_{\bx}\cdot\left(\bSigma\bs\right) + \frac{1}{2}\left\|\bSigma^{\frac{1}{2}}\bs\right\|^2 + \frac{1}{2}\langle\bx,\bs \rangle + \frac{d}{2},\\
q_0(\bx) &= \log\left(p_0(\bx)\right) ,
\end{align}
where the exact LL also requires Monte Carlo simulation.
In sum, the exact score function, LL, and PDF cannot be analytically obtained, requiring Monte Carlo simulation.

\subsection{Geometric Brownian Motion}\label{appendix:gbm}
\textbf{SDE and exact solution}. We consider the geometric Brownian motion and an anisotropic Log-normal as the initial condition:
\begin{align}
d\bx = \frac{1}{2}\exp(-t)\bx dt + \exp(-t/2)\operatorname{diag}(\bx) d\bw_t. \quad p_0(\bx) \sim \text{Log-Normal}(0, \bSigma).
\end{align}
where $\operatorname{diag}(\bx)$ returns a matrix with the input vector $\bx$ as the diagonal.
The exact solution is also a Log-normal $p_t(\bx) \sim \text{Log-Normal}(0, \boldsymbol{\Sigma}_t)$ where $\bSigma_t = \bSigma + (1 - \exp(-t))\bI$ and the exact score function is $\bs_t(\bx) = -\frac{1}{\bx} - \bSigma_t^{-1}\frac{\log\bx}{\bx}$. Since the geometric Brownian motion is more complex, we validate the LL ODE only, as the correctness of the score PDE can be directly inferred from the LL ODE by taking the gradient with respect to $\bx$.

\noindent\textbf{Validation of the LL ODE}. The exact LL is given as follows.
\begin{align}
q_t(\bx) = -\frac{d}{2}\log(2\pi) - \frac{1}{2}\log\operatorname{det}(\bSigma_t) - \frac{1}{2}(\log\bx)^\mathrm{T}\bSigma_t^{-1}(\log\bx) - \sum_{i=1}^d \log\bx_i.
\end{align}
The LL ODE is given by
\begin{align}
\partial_t q_t(\bx) = \frac{\exp(-t)}{2}\nabla_{\bx}\cdot\left(\operatorname{diag}(\bx)\operatorname{diag}(\bx)\bs\right) + \frac{\exp(-t)}{2}\left\|\operatorname{diag}(\bx)\bs\right\|^2 - \langle\bA,\bs \rangle - \nabla\cdot\bA.
\end{align}
On the left-hand side of the LL ODE,
\begin{align}
&\partial_t q_t(\bx) = -\frac{\exp(-t)}{2}\operatorname{Tr}\left(\bSigma_t^{-1}\right) + \frac{\exp(-t)}{2}(\log\bx)^\mathrm{T}\bSigma^{-1}_t\bSigma^{-1}_t(\log\bx).
\end{align}
For the first term on the right-hand side, we denote $\boldsymbol{1}$ as an all-one vector in $\mathbb{R}^d$, then
\begin{equation}
\begin{aligned}
&\quad\frac{1}{2}\nabla_{\bx}\cdot\left(\operatorname{diag}(\bx)\operatorname{diag}(\bx)\bs\right) = -\frac{1}{2}\nabla_{\bx}\cdot\left(\operatorname{diag}(\bx)\left(\boldsymbol{1} + \bSigma_t^{-1}\log\bx\right)\right) \\
&= -\frac{d}{2} - \frac{1}{2}\nabla_{\bx}\cdot\left(\operatorname{diag}(\bx)\bSigma_t^{-1}\log\bx\right) = -\frac{d}{2}  - \frac{1}{2}\boldsymbol{1}^\mathrm{T}\bSigma_t^{-1}\log\bx -\frac{1}{2} \operatorname{Tr}\left(\bSigma_t^{-1}\right).
\end{aligned}
\end{equation}
For the second term
\begin{equation}
\begin{aligned}
&\quad\frac{1}{2}\left\|\operatorname{diag}(\bx)\bs\right\|^2 = \frac{1}{2}\left\|\boldsymbol{1} + \bSigma_t^{-1}\log\bx\right\|^2 = \frac{1}{2}\left(\boldsymbol{1}^\mathrm{T} + (\log\bx)^\mathrm{T}\bSigma_t^{-1}\right)\left(\boldsymbol{1} + \bSigma_t^{-1}\log\bx\right)\\
&=\frac{d}{2}+ \frac{1}{2}(\log\bx)^\mathrm{T}\bSigma^{-1}_t \bSigma^{-1}_t(\log\bx) + \boldsymbol{1}^\mathrm{T}\bSigma^{-1}_t\log\bx
\end{aligned}
\end{equation}
According to equation (\ref{eq:score_pde}), the quantity $\bdf = \frac{1}{2}\exp(-t)\bx$ and $\bG = \exp(-t/2)\operatorname{diag}(\bx) $. Hence,
\begin{align}
&\bA = \bdf - \frac{1}{2}\nabla\cdot[\bG \bG^\mathrm{T}] = \frac{1}{2}\exp(-t)\bx - \exp(-t)\bx = -\frac{1}{2}\exp(-t)\bx.
\end{align}
\begin{align}
 \nabla \cdot \bA = d\left( - \frac{1}{2}\exp(-t)\right), \quad \langle\bA,\bs \rangle = \frac{1}{2} \exp(-t)\left(d + \boldsymbol{1}^\mathrm{T} \bSigma_t^{-1}\log\bx\right)
\end{align}
Taking everything together, we validate the solution.

\section{Failure of Monte Carlo Simulation in High Dimensional PDF}\label{appendix:mc_failure}
\subsection{Gaussian Case}\label{appendix:mc}
This appendix demonstrates the failure of vanilla Monte Carlo simulation in high dimensional PDF, where the convolution of two Gaussian is considered to test the Monte Carlo performance. Specifically, it is well-known that
\begin{equation}
\int_{\mathbb{R}^d} p_\mathcal{N}(\by; 0, \bI) p_\mathcal{N}(\bx - \by; 0, \bI)d\by = p_\mathcal{N}(\bx; 0, 2\bI),
\end{equation}
where $p_{\mathcal{N}}(\bx; \boldsymbol{\mu}, \bSigma) = \frac{1}{(2\pi)^{d/2}\operatorname{det}(\bSigma)^{1/2}}\exp\left(-\frac{(\bx - \boldsymbol{\mu})^\mathrm{T} \bSigma^{-1} (\bx - \boldsymbol{\mu})}{2}\right)$ is the Gaussian PDF with mean $\boldsymbol{\mu} \in \mathbb{R}^d$ and covariance matrix $\bSigma \in \mathbb{R}^{d \times d}$. The integral demonstrates the sum of two unit-Gaussian is another Gaussian $\mathcal{N}(0, 2\bI)$. We simulate the integral via Monte Carlo to test its accuracy. 
Specifically, we sample 100 test points from $\mathcal{N}(0, 2\bI)$ and use $M = 10^7$ samples for Monte Carlo simulation, i.e., for all test points $\bx \sim \mathcal{N}(0, 2\bI)$,
\begin{equation}
\begin{aligned}
\int_{\mathbb{R}^d} p_\mathcal{N}(\by; 0, \bI) p_\mathcal{N}(\bx - \by; 0, \bI)d\by &= \mathbb{E}_{\by \sim \mathcal{N}(0, \bI)} \left[p_\mathcal{N}(\bx - \by; 0, \bI)\right] \\
&\approx \frac{1}{M}\sum_{m=1}^M p_\mathcal{N}(\bx - \by_m; 0, \bI),\quad \by_m \sim \mathcal{N}(0,\bI).
\end{aligned}
\end{equation}
As we mentioned before, the PDF value of the target distribution $p_\mathcal{N}(\bx; 0, 2\bI)$ can be even smaller than the computer simulation accuracy. Thus, directly estimating the PDF will generate zero values. Hence, we compute the LL adopt the following ``normalize and sum" procedure:
\begin{enumerate}
\item Fix the test point $\bx$, compute the LL $q_m = \log p_\mathcal{N}(\bx - \by_m; 0, \bI)$, where $\by_m \sim \mathcal{N}(0, \bI)$. Without loss of generality, we assume the Monte Carlo samples are arranged in descending order, i.e., $q_1 \geq q_2 \geq \cdots \geq q_M$.
\item Normalize $\Tilde{q}_m = q_m - q_1$. Then, we can ensure that $\Tilde{q}_1 = 1 \geq \Tilde{q}_2 \geq \cdots \geq \Tilde{q}_M$. And compute $\frac{1}{M}\sum_{m=1}^M \exp\left(\Tilde{q}_m\right)$. The exact LL can then be estimated via $\log\left[\frac{1}{M}\sum_{m=1}^M \exp\left(\Tilde{q}_m\right)\right] + q_1$. The estimated PDF can also be derived from the estimated LL. 
\end{enumerate}
Our above procedure avoids numerical instability and exactly computes the desired Monte Carlo simulation.
We use NumPy \cite{harris2020array}, one of the most widely used packages for scientific computing and scientific machine learning, for the Monte Carlo simulation.  
New developments in \cite{VonoDC22} on sampling may be used but may not address small scales of the probability density function. A density estimator using neural networks \cite{kobyzev2020normalizing,LiuXJW21} can be an alternative approach while we are not aware of 
any work addressing the issues of small scales of high dimensional PDFs. 
%
The results for Monte Carlo simulation in 10D, 20D, 30D, and 50D are presented in Table \ref{tab:mc}, where we show the relative $L_2$ and $L_\infty$ error for the LL and PDF.
\begin{table}[htbp]
\centering
\begin{tabular}{|c|c|c|c|c|}
\hline
Dimension & 10 & 20 & 30 & 50 \\ \hline
LL $L_2$ Error & 4.739E-4 & 1.121E-3 & 6.071E-3 & 1.634E-2 \\ \hline
LL $L_\infty$ Error & 8.912E-4 & 2.047E-3 & 1.162E-2 & 2.556E-2 \\ \hline
PDF $L_2$ Error & 1.558E-3 & 1.006E-2 & 1.217E-1 & 4.390E-1 \\ \hline
PDF $L_\infty$ Error & 1.712E-3 & 1.072E-2 & 1.131E-1 & 3.618E-1 \\ \hline
\end{tabular}
\caption{Results for vanilla Monte Carlo Simulation for Gaussian}
\label{tab:mc}
\end{table}

\begin{figure}
    \centering
    \includegraphics[scale=0.24]{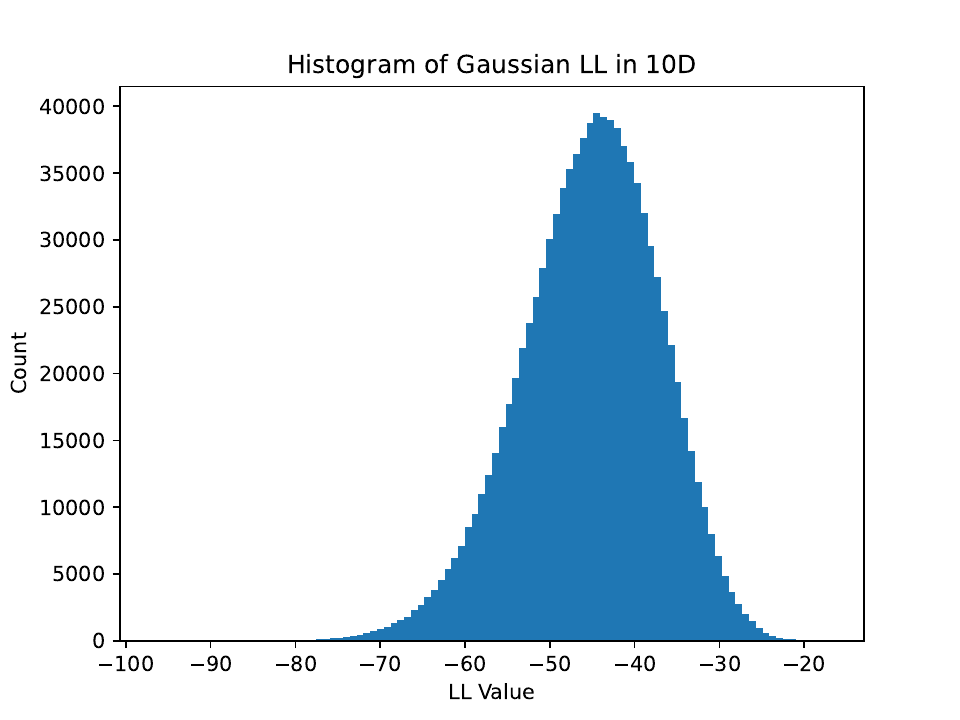}
    \includegraphics[scale=0.24]{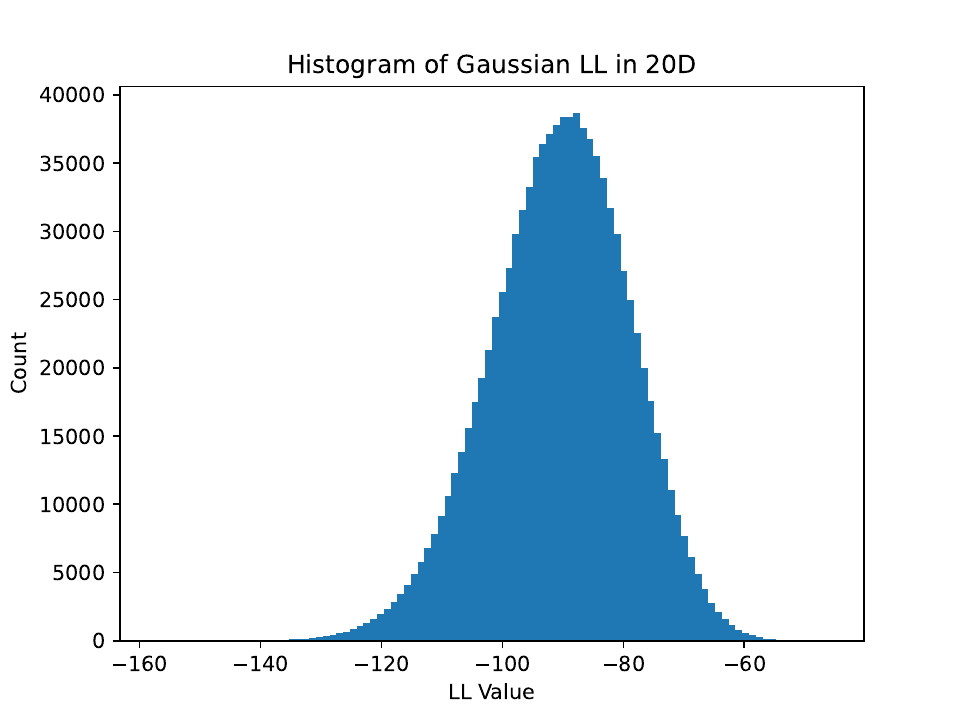}
    \includegraphics[scale=0.24]{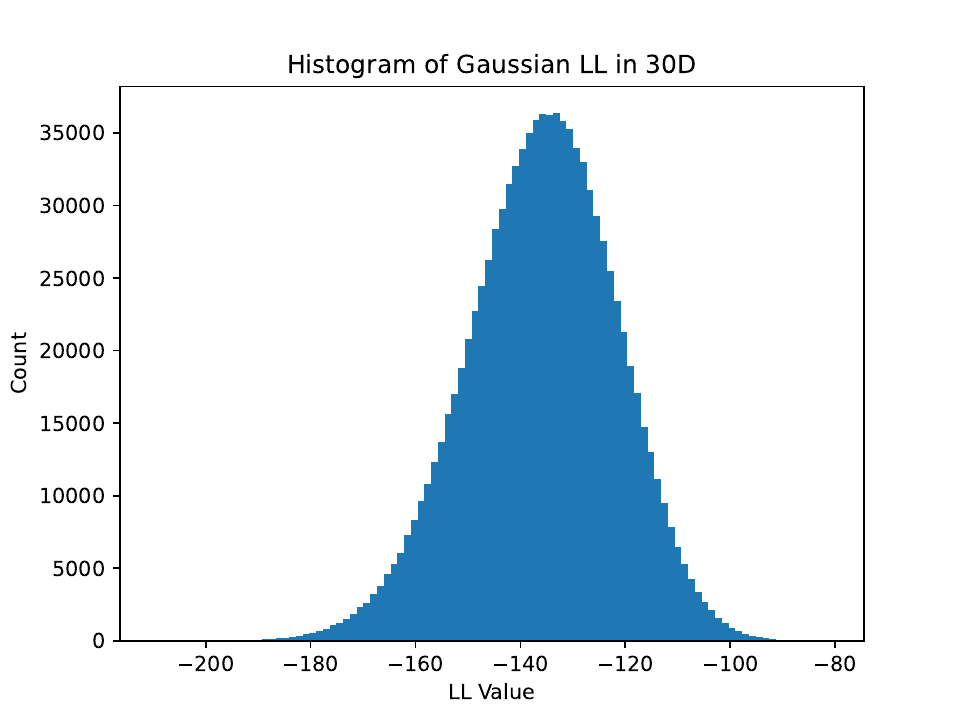}
    \includegraphics[scale=0.24]{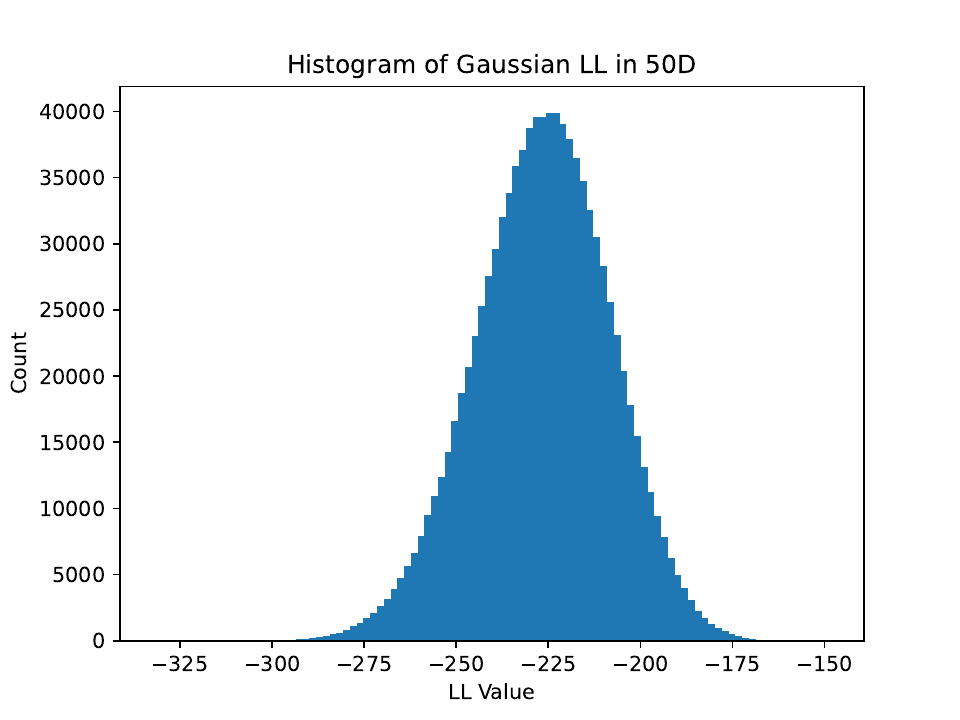}
    \caption{Histrogram of the $\log p_\mathcal{N}(\bx - \by_m; 0, \bI)$ values in various dimensions. The discrepancy between large and small values becomes larger with dimension, causing the numerical issue when summing the PDF value.}
    \label{fig:gaussian_mc}
\end{figure}
Due to the numerical instability, the error of the Monte Carlo simulation grows with the dimensionality. Consequently, our experiment tests our score-based SDE solver in less than 30D cases; otherwise, the Monte Carlo simulation generated reference will not be reliable. The error comes from two sources. 
\begin{itemize}
\item The inherent error of Monte Carlo simulation.
\item The numerical instability. Since the PDF values $p_\mathcal{N}(\bx - \by_m; 0, \bI)$ for different $m$ differs to a large extent, a lot of small PDF values will be regarded as zero in a computer simulation, although it does contribute to the integral. This phenomenon becomes more severe in higher dimensions since the discrepancy between large and small values in $p_\mathcal{N}(\bx - \by_m; 0, \bI)$ for different $m$ tends to be larger. Discarding too many values leads to substantial error, although these values are small.
\end{itemize}

Next, we theoretically analyze the numerical issue when summing the PDFs during the Monte Carlo simulation. Consider the PDF value of a unit Gaussian $\mathcal{N}(0, \bI)$; it is a chi-square distribution with freedom $d$ where $d$ is the dimensionality of the Gaussian, whose mean is $d$ and variance is $2d$. Thus, the discrepancy between large and small values in Gaussian PDF becomes linearly larger with dimensionality, leading to more significant numerical error during vanilla Monte Carlo simulation. This observation is validated in Figure \ref{fig:gaussian_mc}, where we show the histogram of the $\log p_\mathcal{N}(\bx - \by_m; 0, \bI)$ values for a fixed $\bx$ in various dimensions. For instance, in the 50D case, the $\log p_\mathcal{N}(\bx - \by_m; 0, \bI)$ spans from -275 to -175, which means the PDF spans from $\exp(-275)$ to $\exp(-175)$.

This issue can actually be addressed through careful handling of the PDF, e.g., extending the precision of float numbers in computer simulation. However, increasing the precision of floating-point numbers implies greater memory consumption and slower computations, which are also incompatible with current packages and programming languages like NumPy and Python, not to mention failure to fast sampling due to the massive sampling in Monte Carlo.

\subsection{Log-Normal Case}\label{appendix:mc2}
This appendix illustrates the Monte Carlo simulation's substantial errors in the Log-normal distribution's PDF context. Specifically, in the previous Appendix \ref{appendix:mc}, we demonstrated that the sum of two Gaussians is Gaussian, a common occurrence in Brownian motion. Here, we show that the product of two Log-normals is also Log-normal, corresponding to geometric Brownian motion. We denote the PDF of Log-normal with mean $\boldsymbol{\mu}$ and variance $\bSigma$ as 
$
p_{\log\mathcal{N}}(\bx; \boldsymbol{\mu}, \bSigma) = \frac{1}{(2\pi)^{d/2}\operatorname{det}(\bSigma)^{1/2}}\left(\prod_{i=1}^d\frac{1}{\bx_i}\right)\exp\left(-\frac{(\log\bx - \boldsymbol{\mu})^\mathrm{T} \bSigma^{-1} (\log\bx - \boldsymbol{\mu})}{2}\right), 
$
then
\begin{equation}
\int_{\mathbb{R}^d} p_{\log\mathcal{N}}(\by; 0, 0.1\bI) p_{\log\mathcal{N}}(\bx / \by; 0, \bI)d\by = p_{\log\mathcal{N}}(\bx; 0, 1.1\bI).
\end{equation}
We test the Monte Carlo simulation using the equality, where we denote the Log-normal as $\log\mathcal{N}$:
\begin{equation*}
\int_{\mathbb{R}^d} p_{\log\mathcal{N}}(\by; 0, 0.1\bI) p_{\log\mathcal{N}}(\bx / \by; 0, \bI)d\by  = \mathbb{E}_{\by \sim \log\mathcal{N}(0, 0.1\bI)} \left[p_{\log\mathcal{N}}(\bx / \by; 0, \bI)\right] \approx \frac{1}{M}\sum_{m=1}^M p_{\log\mathcal{N}}(\bx / \by_m; 0, \bI),
\end{equation*}
where $\by_m \sim \log\mathcal{N}(0,0.1\bI)$. Specifically, we sample 100 test points from $\log\mathcal{N}(0, 1.1\bI)$ and use $M = 10^7$ samples for Monte Carlo simulation, i.e., for all test points $\bx$. We also use the ``normalize and sum" trick to ensure numerical stability. The results for Monte Carlo simulation in 3D, 10D, 20D, and 30D are presented in Table \ref{tab:mc2}, where we show the relative $L_2$ and $L_\infty$ error for the LL and PDF. 
\begin{table}[htbp]
\centering
\begin{tabular}{|c|c|c|c|c|}
\hline
Dimension & 3 & 10 & 20 & 30 \\ \hline
LL $L_2$ Error & 4.615E-2 & 3.993E-2 & 3.737E-2 & 3.535E-2 \\ \hline
LL $L_\infty$ Error & 3.992E-2 & 4.676E-2 & 4.418E-2 & 4.441E-2 \\ \hline
PDF $L_2$ Error & 1.794E-1 & 1.166E-1 & 2.667E-1 & 6.414E-1 \\ \hline
PDF $L_\infty$ Error & 1.468E-1 & 9.528E-2 & 2.693E-1 & 6.411E-1 \\ \hline
\end{tabular}
\caption{Results for Monte Carlo Simulation for Log-normal}
\label{tab:mc2}
\end{table}
Despite stable and low LL errors, the Monte Carlo simulation exhibits significant errors across all dimensions in PDF errors due to the heavy tail of the Log-normal distribution. Previously, we used Gaussian for Monte Carlo, which has a short tail, enabling more accurate results.

It is worth noting that due to the long tail of the Log-normal distribution, the PDF convolution integral is challenging to compute even in low dimensions. Gaussian quadrature methods such as Gauss-Hermite quadrature and Gauss-Laguerre fail in this context because they assume the integrand decays exponentially.

\textbf{Long tail leads to substantial PDF error}. Long tail means many rare events; thus, we have many samples with small PDF values. These large numbers of small PDF values will be truncated during vanilla Monte Carlo, leading to Monte Carlo's substantial error.

\textbf{Relationship between PDF error and LL error}. 
LL is linear, while PDF is exponential concerning the dimension. Hence, a small LL error can not guarantee a small PDF error. For example, if the exact LL is -100, while the predicted LL is -99, Then the relative LL error will be 1\%. However, the relative PDF error is $|\exp(-99) - \exp(-100)| / |\exp(-100)| = e - 1$, which is more than 200\%. Hence, PDF relative error testing becomes unstable in high dimensions.

\subsection{Cauchy Case}\label{appendix:mc3}
To ascertain the inadequacy of Monte Carlo in handling long-tail distributions, we conducted additional validation using the Cauchy distribution. The sum of two isotropic Cauchy distributions is also a Cauchy distribution. By denoted the Cauchy PDF with coefficient $\gamma$ as $\mathcal{C}(\gamma)$ with the PDF
$
p_\mathcal{C}(\bx;\gamma) = \frac{\Gamma\left(\frac{1 + d}{2}\right)}{\Gamma\left(\frac{1}{2}\right)\pi^{\frac{d}{2}}\gamma^{\frac{d}{2}}\left[1 + \frac{\Vert\bx\Vert^2}{\gamma^2}\right]^{\frac{d+1}{2}}}
$,
then
\begin{equation}
\int_{\mathbb{R}^d} p_{\mathcal{C}}(\by; 1) p_{\mathcal{C}}(\bx - \by; 1)d\by = p_{\mathcal{C}}(\bx; 2).
\end{equation}
We test the Monte Carlo simulation using the equality:
\begin{equation*}
\int_{\mathbb{R}^d} p_{\mathcal{C}}(\by; 1) p_{\mathcal{C}}(\bx - \by; 1)d\by  = \mathbb{E}_{\by \sim \mathcal{C}(1)} \left[p_{\mathcal{C}}(\bx - \by; 1)\right] \approx \frac{1}{M}\sum_{m=1}^M p_{\mathcal{C}}(\bx - \by_m; 1),
\end{equation*}
where $\by_m \sim \mathcal{C}(1)$. Specifically, we sample 100 test points from $\mathcal{C}(2)$ and use $M = 10^7$ samples for Monte Carlo simulation, i.e., for all test points $\bx$. We also use the ``normalize and sum" trick to ensure numerical stability. The results for Monte Carlo simulation in 3D, 5D, and 10D are presented in Table \ref{tab:mc3}.
\begin{table}[htbp]
\centering
\begin{tabular}{|c|c|c|c|}
\hline
Dimension & 3 & 5 & 10 \\ \hline
LL $L_2$ Error & 2.781E-2 & 1.818E-2 & 1.307E-2 \\ \hline
LL $L_\infty$ Error & 2.729E-2 & 2.717E-2 & 1.777E-2 \\ \hline
PDF $L_2$ Error & 1.297E-1 & 9.034E-1 & 6.148E0 \\ \hline
PDF $L_\infty$ Error & 1.731E-1 & 1.350E0 & 6.558E0 \\ \hline
\end{tabular}
\caption{Results for vanilla Monte Carlo Simulation for the long-tail Cauchy distribution}
\label{tab:mc3}
\end{table}
Similar to the long-tailed Log-normal distribution, Monte Carlo simulations also exhibit significant PDF errors in the estimation of the Cauchy distribution convolution. This is attributed to the numerous small PDF values in the long-tailed distribution being ignored or truncated due to the limited precision of computer simulations, leading to an accumulation of errors.

\newpage
\bibliographystyle{plain}
\bibliography{references}

\begin{thebibliography}{10}

\bibitem{Beck2019DeepSM}
Christian Beck, Sebastian Becker, Patrick Cheridito, Arnulf Jentzen, and Ariel Neufeld.
\newblock Deep splitting method for parabolic {PDEs}.
\newblock {\em SIAM J. Sci. Comput.}, 43:A3135--A3154, 2019.

\bibitem{beck2020overcoming}
Christian Beck, Lukas Gonon, and Arnulf Jentzen.
\newblock Overcoming the curse of dimensionality in the numerical approximation of high-dimensional semilinear elliptic partial differential equations.
\newblock {\em arXiv preprint arXiv:2003.00596}, 2020.

\bibitem{jax2018github}
James Bradbury, Roy Frostig, Peter Hawkins, Matthew~James Johnson, Chris Leary, Dougal Maclaurin, George Necula, Adam Paszke, Jake Vander{P}las, Skye Wanderman-{M}ilne, and Qiao Zhang.
\newblock {JAX}: composable transformations of {P}ython+{N}um{P}y programs, 2018.

\bibitem{chen2021solving}
Xiaoli Chen, Liu Yang, Jinqiao Duan, and George~Em Karniadakis.
\newblock Solving inverse stochastic problems from discrete particle observations using the {Fokker--Planck} equation and physics-informed neural networks.
\newblock {\em SIAM Journal on Scientific Computing}, 43(3):B811--B830, 2021.

\bibitem{deng2009finite}
Weihua Deng.
\newblock Finite element method for the space and time fractional {Fokker--Planck} equation.
\newblock {\em SIAM journal on numerical analysis}, 47(1):204--226, 2009.

\bibitem{feng2021solving}
Xiaodong Feng, Li~Zeng, and Tao Zhou.
\newblock Solving time dependent {Fokker-Planck} equations via temporal normalizing flow.
\newblock {\em arXiv preprint arXiv:2112.14012}, 2021.

\bibitem{guo2022normalizing}
Ling Guo, Hao Wu, and Tao Zhou.
\newblock Normalizing field flows: Solving forward and inverse stochastic differential equations using physics-informed flow models.
\newblock {\em Journal of Computational Physics}, 461:111202, 2022.

\bibitem{han2018solving}
Jiequn Han, Arnulf Jentzen, and Weinan E.
\newblock Solving high-dimensional partial differential equations using deep learning.
\newblock {\em Proceedings of the National Academy of Sciences}, 115(34):8505--8510, 2018.

\bibitem{harris2020array}
Charles~R. Harris, K.~Jarrod Millman, St{\'{e}}fan~J. van~der Walt, Ralf Gommers, Pauli Virtanen, David Cournapeau, Eric Wieser, Julian Taylor, Sebastian Berg, Nathaniel~J. Smith, Robert Kern, Matti Picus, Stephan Hoyer, Marten~H. van Kerkwijk, Matthew Brett, Allan Haldane, Jaime~Fern{\'{a}}ndez del R{\'{i}}o, Mark Wiebe, Pearu Peterson, Pierre G{\'{e}}rard-Marchant, Kevin Sheppard, Tyler Reddy, Warren Weckesser, Hameer Abbasi, Christoph Gohlke, and Travis~E. Oliphant.
\newblock Array programming with {NumPy}.
\newblock {\em Nature}, 585(7825):357--362, September 2020.

\bibitem{he2023learning}
Di~He, Shanda Li, Wenlei Shi, Xiaotian Gao, Jia Zhang, Jiang Bian, Liwei Wang, and Tie-Yan Liu.
\newblock Learning physics-informed neural networks without stacked back-propagation.
\newblock In {\em International Conference on Artificial Intelligence and Statistics}, pages 3034--3047. PMLR, 2023.

\bibitem{hu2023hutchinson}
Zheyuan Hu, Zekun Shi, George~Em Karniadakis, and Kenji Kawaguchi.
\newblock Hutchinson trace estimation for high-dimensional and high-order physics-informed neural networks.
\newblock {\em arXiv preprint arXiv:2312.14499}, 2023.

\bibitem{hu2023tackling}
Zheyuan Hu, Khemraj Shukla, George~Em Karniadakis, and Kenji Kawaguchi.
\newblock Tackling the curse of dimensionality with physics-informed neural networks.
\newblock {\em arXiv preprint arXiv:2307.12306}, 2023.

\bibitem{hu2023bias}
Zheyuan Hu, Zhouhao Yang, Yezhen Wang, George~Em Karniadakis, and Kenji Kawaguchi.
\newblock Bias-variance trade-off in physics-informed neural networks with randomized smoothing for high-dimensional {PDEs}.
\newblock {\em arXiv preprint arXiv:2311.15283}, 2023.

\bibitem{hutzenthaler2021multilevel}
Martin Hutzenthaler, Arnulf Jentzen, Thomas Kruse, et~al.
\newblock Multilevel picard iterations for solving smooth semilinear parabolic heat equations.
\newblock {\em Partial Differential Equations and Applications}, 2(6):1--31, 2021.

\bibitem{hyvarinen2005estimation}
Aapo Hyv{\"a}rinen and Peter Dayan.
\newblock Estimation of non-normalized statistical models by score matching.
\newblock {\em Journal of Machine Learning Research}, 6(4), 2005.

\bibitem{kingma2014adam}
Diederik~P Kingma and Jimmy Ba.
\newblock Adam: A method for stochastic optimization.
\newblock {\em arXiv preprint arXiv:1412.6980}, 2014.

\bibitem{kobyzev2020normalizing}
Ivan Kobyzev, Simon~JD Prince, and Marcus~A Brubaker.
\newblock Normalizing flows: An introduction and review of current methods.
\newblock {\em IEEE transactions on pattern analysis and machine intelligence}, 43(11):3964--3979, 2020.

\bibitem{lai2022regularizing}
Chieh-Hsin Lai, Yuhta Takida, Naoki Murata, Toshimitsu Uesaka, Yuki Mitsufuji, and Stefano Ermon.
\newblock Regularizing score-based models with score {Fokker-Planck} equations.
\newblock {\em arXiv preprint arXiv:2210.04296}, 2022.

\bibitem{LiuXJW21}
Qiao Liu, Jiaze Xu, Rui Jiang, and Wing~Hung Wong.
\newblock Density estimation using deep generative neural networks.
\newblock {\em Proceedings of the National Academy of Sciences}, 118(15):e2101344118, 2021.

\bibitem{lu2021physics}
Lu~Lu, Raphael Pestourie, Wenjie Yao, Zhicheng Wang, Francesc Verdugo, and Steven~G Johnson.
\newblock Physics-informed neural networks with hard constraints for inverse design.
\newblock {\em SIAM Journal on Scientific Computing}, 43(6):B1105--B1132, 2021.

\bibitem{lu2022learning}
Yubin Lu, Romit Maulik, Ting Gao, Felix Dietrich, Ioannis~G Kevrekidis, and Jinqiao Duan.
\newblock Learning the temporal evolution of multivariate densities via normalizing flows.
\newblock {\em Chaos: An Interdisciplinary Journal of Nonlinear Science}, 32(3), 2022.

\bibitem{oksendal2013stochastic}
Bernt Oksendal.
\newblock {\em Stochastic differential equations: an introduction with applications}.
\newblock Springer Science \& Business Media, 2013.

\bibitem{pang2020efficient}
Tianyu Pang, Kun Xu, Chongxuan Li, Yang Song, Stefano Ermon, and Jun Zhu.
\newblock Efficient learning of generative models via finite-difference score matching.
\newblock {\em Advances in Neural Information Processing Systems}, 33:19175--19188, 2020.

\bibitem{penwarden2023unified}
Michael Penwarden, Ameya~D Jagtap, Shandian Zhe, George~Em Karniadakis, and Robert~M Kirby.
\newblock A unified scalable framework for causal sweeping strategies for physics-informed neural networks (pinns) and their temporal decompositions.
\newblock {\em arXiv preprint arXiv:2302.14227}, 2023.

\bibitem{raissi2019physics}
Maziar Raissi, Paris Perdikaris, and George~E Karniadakis.
\newblock Physics-informed neural networks: A deep learning framework for solving forward and inverse problems involving nonlinear partial differential equations.
\newblock {\em Journal of Computational Physics}, 378:686--707, 2019.

\bibitem{sepehrian2015numerical}
Behnam Sepehrian and Marzieh~Karimi Radpoor.
\newblock Numerical solution of non-linear {Fokker--Planck} equation using finite differences method and the cubic spline functions.
\newblock {\em Applied mathematics and computation}, 262:187--190, 2015.

\bibitem{song2019sliced}
Yang Song, Sahaj Garg, Jiaxin Shi, and Stefano Ermon.
\newblock Sliced score matching: {A} scalable approach to density and score estimation.
\newblock In {\em Proceedings of the Thirty-Fifth Conference on Uncertainty in Artificial Intelligence, {UAI} 2019, Tel Aviv, Israel, July 22-25, 2019}, page 204, 2019.

\bibitem{song2021scorebased}
Yang Song, Jascha Sohl-Dickstein, Diederik~P Kingma, Abhishek Kumar, Stefano Ermon, and Ben Poole.
\newblock Score-based generative modeling through stochastic differential equations.
\newblock In {\em International Conference on Learning Representations}, 2021.

\bibitem{tang2022adaptive}
Kejun Tang, Xiaoliang Wan, and Qifeng Liao.
\newblock Adaptive deep density approximation for {Fokker-Planck} equations.
\newblock {\em Journal of Computational Physics}, 457:111080, 2022.

\bibitem{VonoDC22}
Maxime Vono, Nicolas Dobigeon, and Pierre Chainais.
\newblock High-dimensional gaussian sampling: A review and a unifying approach based on a stochastic proximal point algorithm.
\newblock {\em SIAM Review}, 64(1):3--56, 2022.

\bibitem{wang20222}
Chuwei Wang, Shanda Li, Di~He, and Liwei Wang.
\newblock Is \$l{\textasciicircum}2\$ physics informed loss always suitable for training physics informed neural network?
\newblock In Alice~H. Oh, Alekh Agarwal, Danielle Belgrave, and Kyunghyun Cho, editors, {\em Advances in Neural Information Processing Systems}, 2022.

\bibitem{zhai2022deep}
Jiayu Zhai, Matthew Dobson, and Yao Li.
\newblock A deep learning method for solving {Fokker-Planck} equations.
\newblock In {\em Mathematical and scientific machine learning}, pages 568--597. PMLR, 2022.

\bibitem{zhang2022solving}
Hao Zhang, Yong Xu, Qi~Liu, Xiaolong Wang, and Yongge Li.
\newblock Solving {Fokker--Planck} equations using deep kd-tree with a small amount of data.
\newblock {\em Nonlinear Dynamics}, 108(4):4029--4043, 2022.

\bibitem{zhao2023tensor}
Yequan Zhao, Xinling Yu, Zhixiong Chen, Ziyue Liu, Sijia Liu, and Zheng Zhang.
\newblock Tensor-compressed back-propagation-free training for (physics-informed) neural networks.
\newblock {\em arXiv preprint arXiv:2308.09858}, 2023.

\end{thebibliography}
\end{document}